\documentclass{article}
\usepackage{iclr2027_conference,times}
\iclrfinalcopy

\usepackage{amsmath,amsfonts,bm}

\def\eqref#1{equation~\ref{#1}}

\def\1{\bm{1}}

\DeclareMathAlphabet{\mathsfit}{\encodingdefault}{\sfdefault}{m}{sl}
\SetMathAlphabet{\mathsfit}{bold}{\encodingdefault}{\sfdefault}{bx}{n}

\usepackage{graphicx}
\usepackage{booktabs}
\usepackage{amsmath,amssymb}
\usepackage{multirow}
\usepackage{hyperref}
\usepackage{url}
\hypersetup{
  hidelinks,
  pdftitle={Knowing When to Stop Thinking: A Dissection of Expected Free Energy as a Test-Time Compute Controller for Latent World-Model Planning},
  pdfauthor={Tong Zhang, Zhiyuan Shi, Yun Peng, Tao Xie},
  pdfsubject={Preprint}
}
\usepackage{subcaption}
\usepackage{xcolor}

\newcommand{\deltaefe}{\Delta\mathrm{EFE}}
\newcommand{\reviewcomment}[1]{}

\title{Knowing When to Stop Thinking\\ A Dissection of Expected Free Energy as a\\ Test-Time Compute Controller for\\ Latent World-Model Planning}

\author{Tong Zhang\textsuperscript{1} \quad Zhiyuan Shi \quad
Yun Peng\textsuperscript{1} \quad Tao Xie\textsuperscript{2}\\
\textsuperscript{1}Fudan University \qquad
\textsuperscript{2}Peking University}

\begin{document}
\maketitle
\pagestyle{plain}

\begin{abstract}
A planner using a learned world model can improve its predicted outcome while choosing a worse action. We study when expected free energy (EFE) can help decide how much computation to spend on a plan. An EFE-inspired objective combines goal distance with disagreement among stochastic predictions. We test candidate scoring, stopping, and online predictor updates separately. Stopping when elite cost stalls reduces refinement on tasks whose success plateaus with depth, with mean success close to fixed-depth references. At the larger tested weights, rewarding disagreement lowers success with a frozen model. Penalizing it helps at some task and weight settings but harms others. When training data omit a region needed to reach the goal, online updates recover part of the resulting loss. The tested cost-stall rule can fail when further optimization exploits model errors, because predicted cost keeps falling after task success peaks. Cost progress can guide computation on a plateau, but protecting performance after a peak requires information about the consequences of the chosen actions.
\end{abstract}

\section{Introduction}
A robot planning from images can improve its next action by repeatedly imagining and comparing possible futures. Each refinement step costs computation. Stopping too early may leave a useful action undiscovered. Continuing too long can favor an action that exploits errors in the learned dynamics. The planner must decide whether further optimization will improve the action it executes.

Test-time computation has improved iterative reasoning and model-based control. Recurrent language models refine internal representations before producing an answer \citep{geiping2025recurrent,wang2025hrm,jolicoeur2025trm}. World-model agents predict future states to evaluate candidate actions \citep{hafner2023dreamerv3,assran2025vjepa2}. A larger optimization budget gives the planner more opportunities to improve a plan. Its objective, however, remains a prediction of task performance. A fixed budget cannot respond to differences between instances, and convergence of the objective does not establish that the predicted consequences are correct.

Expected free energy (EFE) connects goal achievement and information seeking \citep{friston2010fep,parr2022active}. Its pragmatic component favors preferred outcomes, and its epistemic component values information. These components suggest ways to control planning. Goal progress may indicate when refinement has stalled. Prediction uncertainty may change which plan the agent selects. New observations may improve the model before the next decision. Disagreement-based exploration methods collect information during learning \citep{sekar2020plan2explore}, whereas uncertainty penalties discourage plans that rely on unreliable predictions \citep{yu2020mopo}. Whether disagreement helps immediate task execution with a frozen predictor remains unclear. Internal progress may also fail to reveal when more planning becomes harmful.

\begin{figure}[t]
\centering
\includegraphics[width=\linewidth]{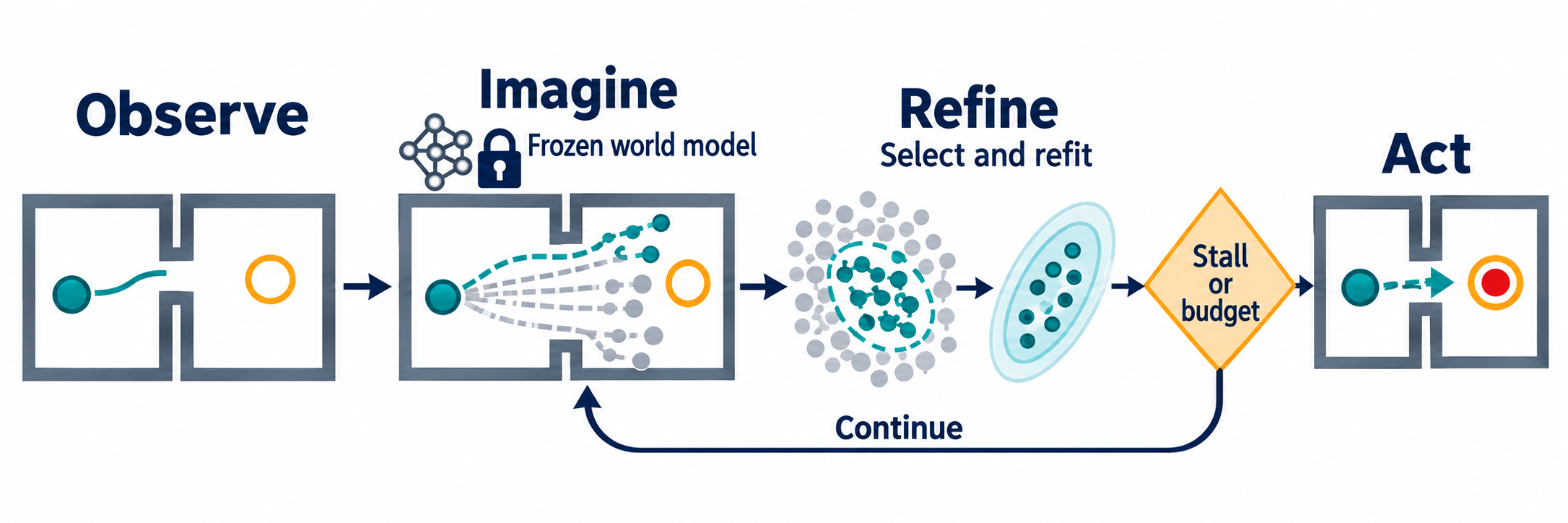}
\par\smallskip
\includegraphics[width=\linewidth]{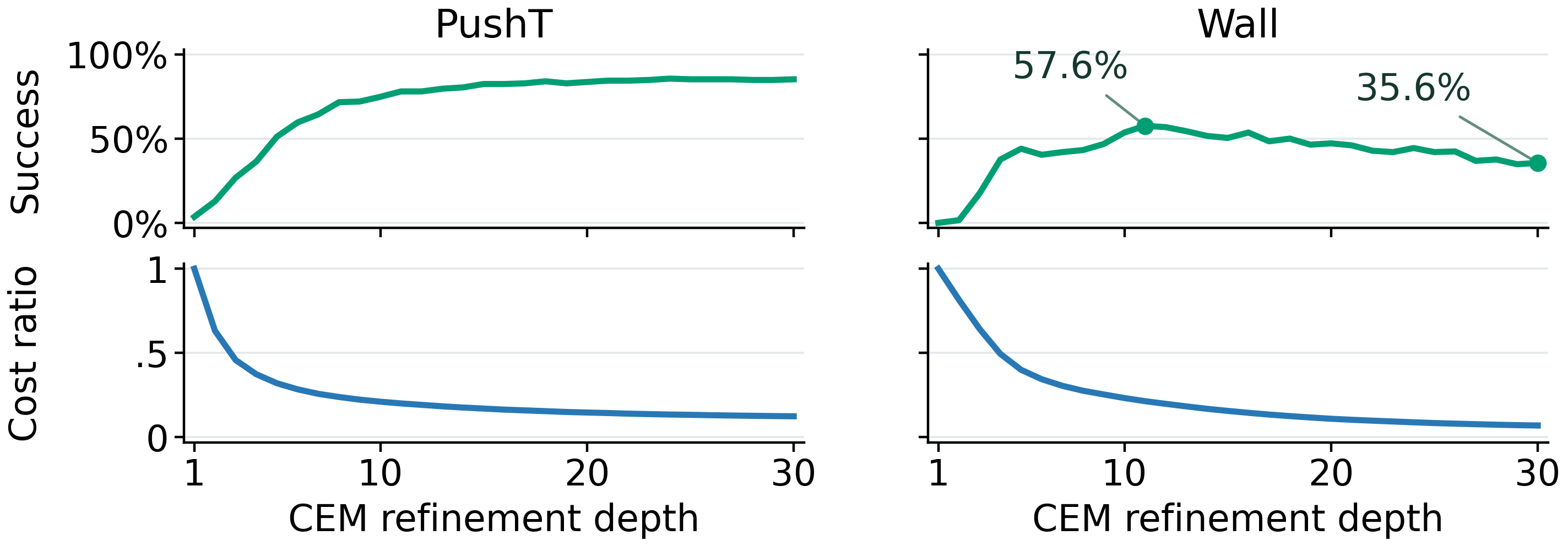}
\caption{Planning and measured effects of refinement. Top, the planning workflow refines candidate actions with a frozen world model until cost progress stalls or the budget is exhausted. Bottom, DINO-WM results from five seeds with 50 starts per seed. Success and model cost are plotted on separate axes. Each cost curve is divided by its first-iteration value within each seed before averaging. PushT success reaches a plateau. Wall success peaks at iteration 11 at $57.6\%$ and falls to $35.6\%$ at iteration 30 as model cost continues to decrease. All plotted curves come from recorded evaluations.}
\label{fig:teaser}
\end{figure}

We test these decisions with a compact latent world model and a sampling planner. In Figure~\ref{fig:teaser}, candidate trajectories provide goal cost and disagreement among stochastic predictions. We vary the disagreement weight and stopping statistic with the model frozen. We then test stopping in additional planner families and remove training data from a required region to study online learning. These comparisons separate the effects of reducing refinement, changing action selection, and updating the predictor.

Cost-stall rules reduce refinement on tasks whose success plateaus with depth, with mean success close to fixed-depth references. A best-elite baseline is competitive and often stops earlier. On tasks whose success declines after a peak, falling cost sustains harmful refinement. Rewarding disagreement reduces frozen-model planning success, whereas penalties help at selected task and weight settings. Online updates improve success when the agent must enter a region missing from its training data.

Our experiments address the stopping decision and the uses of prediction uncertainty.
\begin{itemize}
\item We separate candidate scoring, stopping, and online adaptation in a common world-model planning protocol.
\item We measure when cost-stall controllers reduce planning effort and when model exploitation defeats the tested stopping signals.
\item We measure the task-dependent effects of disagreement penalties and test online learning by removing training data from a required region.
\end{itemize}

\section{Related Work}
Active inference formulates perception and action through free-energy objectives \citep{friston2010fep,parr2022active}. Deep implementations combine learned representations with approximate inference \citep{ueltzhoffer2018daif,fountas2020daif,mazzaglia2021contrastive,mazzaglia2022fep}. The information-seeking term requires assumptions beyond free-energy minimization \citep{millidge2021whence}, and Bellman-optimality results hold under specified conditions \citep{dacosta2023reward}. We use this framework to motivate a cost combining goal distance and stochastic prediction disagreement. Disagreement provides an uncertainty proxy. It is not a calibrated estimate of information gain.

Learned latent dynamics support planning from images \citep{hafner2019planet,hansen2022tdmpc,assran2025vjepa2}. We build on LeWM, a compact joint-embedding predictive architecture that predicts future image representations \citep{maes2026lewm}. Dreamer improves policies through imagined returns \citep{hafner2023dreamerv3}, whereas Plan2Explore uses disagreement to gather experience during learning \citep{sekar2020plan2explore}. Our frozen-model experiments test disagreement for immediate goal-directed execution. The adaptation experiment tests what changes when observed transitions update the predictor.

Adaptive computation allocates different amounts of work to different inputs \citep{graves2016act}. Recurrent and recursive reasoning apply this idea to internal model computation \citep{geiping2025recurrent,wang2025hrm,jolicoeur2025trm}. We study stopping inside sampling-based model predictive control and compare statistics from the same optimization process. Pessimistic offline reinforcement learning addresses unreliable estimates through model uncertainty penalties in MOPO \citep{yu2020mopo} and conservative value estimates in CQL \citep{kumar2020cql}. We compare a disagreement penalty with a disagreement reward. The coverage experiment then tests a setting where the goal requires entering a region absent from training, so avoiding uncertain dynamics cannot solve the task.

\section{Method}
\label{sec:method}
\begin{figure}[t]
\centering
\includegraphics[width=\linewidth]{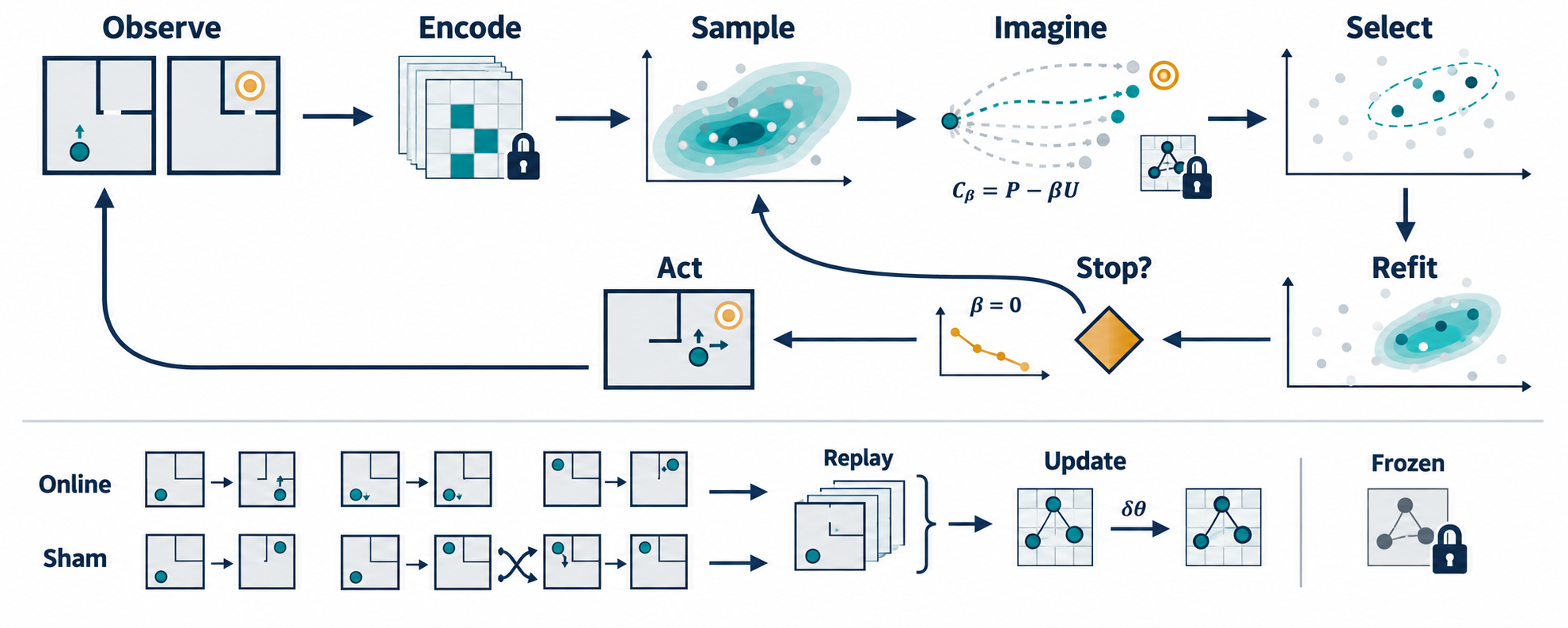}
\caption{Planning and online adaptation. The upper loop encodes images, samples actions, predicts and scores futures, selects elites, and refits the proposal. The stopping rule either continues refinement or executes the selected action. The lower branch updates the predictor using collected transitions, replay, and a parameter anchor. Shuffled-target updates and a frozen predictor provide controls.}
\label{fig:pipeline}
\end{figure}

Figure~\ref{fig:pipeline} shows the planning loop. A fixed world model predicts candidate futures, the candidate cost determines which actions refine the proposal, and a stopping rule determines when to execute an action. The lower branch updates the predictor from observations in a separate experiment. We vary scoring, stopping, and adaptation separately.

\subsection{Candidate Scoring}
The planner compares possible actions before executing them. Let $o_t$ denote the current image, $z_t$ its encoded representation, and $z^g$ the encoded goal image. A latent predictor $f_\theta$ advances the state representation under a candidate action sequence $a$. The cross-entropy method (CEM) \citep{rubinstein1999cem} samples action sequences from a Gaussian distribution, retains the lowest-cost candidates as an elite set, and refits the distribution to that set. Each refinement iteration repeats this sample, predict, and refit procedure.

To expose both goal progress and predictive variation, we evaluate each candidate with $K$ stochastic predictor rollouts. The terminal prediction in rollout $k$ is $z_T^{(k)}(a)$ and their mean is $\bar z_T(a)=K^{-1}\sum_{k=1}^{K}z_T^{(k)}(a)$. Candidate scoring uses
\begin{equation}
 \mathcal C_\beta(a)=P(a)-\beta U(a),\qquad
 P(a)=\|\bar z_T(a)-z^g\|^2,
 \label{eq:efe}
\end{equation}
where $U(a)=K^{-1}\sum_{k=1}^{K}\|z_T^{(k)}(a)-\bar z_T(a)\|^2$ for $K>1$, and $U(a)=0$ for $K=1$. The goal term $P$ sums squared terminal-coordinate errors. The disagreement term $U$ sums coordinatewise variance with divisor $K$. Positive $\beta$ rewards disagreement and negative $\beta$ penalizes it. This sign change tests two uses of the same measurable signal without changing the predictor.

The objective uses goal distance and disagreement to study the pragmatic and epistemic roles of EFE. Disagreement measures stochastic prediction spread. It does not estimate posterior information gain. The weight experiment enables predictor dropout at every weight, including $\beta=0$, so the comparisons share the same stochastic scoring procedure. The deterministic reference disables dropout and uses a single rollout. At $\beta=0$ it recovers terminal goal-distance scoring. We also evaluate magnitude-normalized weights because $P$ and $U$ can have different magnitudes.

Terminal spread is large when stochastic predictions disagree about where a candidate action sequence will end. A reward favors uncertain predicted outcomes, and a penalty favors more consistent predictions. Changing the sign leaves the goal representation and feasible action space unchanged. It tests how disagreement affects action selection without assuming that greater spread means more useful information.

\subsection{Stopping Rule}
A fixed iteration budget spends the same maximum effort on every solve. We instead monitor how much the elite cost improves. Let $\mathcal E_n$ denote the elite set at iteration $n$, and let $e_n$ be its mean candidate cost. The controller monitors an exponential moving average $E_n$ of $e_n$ and computes relative descent
\begin{equation}
 q_n=\frac{E_{n-1}-E_n}{\max(|E_{n-1}|,10^{-8})}.
 \label{eq:halt}
\end{equation}
It stops at the first eligible iteration for which $q_n<\varepsilon$ throughout a window of $p$ consecutive iterations. Eligibility requires both the minimum depth $n_{\min}$ and a complete patience window. The floor in the denominator handles costs near zero. If no stall occurs, refinement ends at the cap $N$.

The patience window prevents an isolated small improvement from ending the solve. Smoothing reduces sensitivity to changes in the sampled elite set, and the minimum depth gives the optimizer time to establish a candidate distribution. A cost increase also satisfies the stall inequality because relative descent becomes negative. The rule detects insufficient recent improvement. It does not prove convergence of the sampling distribution. We compare alternative cost statistics under the same patience and minimum-depth settings.

Write $\hat n$ for the selected stopping iteration. After stopping, the agent executes the first action block, observes the next image, and plans again. Depth measures computation within a control decision. The primary controller uses $\beta=0$ with shared settings across the LeWM tasks, so its descent signal measures improvement in predicted goal distance. We compare it with best-elite cost stall, mean-change, variance-stall, and elite-set stability. Unsmoothed and disagreement-aware variants test changes to the monitored statistic. Patience longer than the available window recovers fixed-depth refinement.

\subsection{Online Adaptation}
Stopping and candidate scoring operate within the current model. Online adaptation changes that model. To test its value, we remove training episodes that enter a specified corner of TwoRoom and evaluate goals inside that region. This creates a coverage hole that goal-directed behavior must encounter. A volume-matched control removes the same amount of data at random. A separate checkpoint matches the hole model's frozen success, distinguishing low initial performance from the location of missing training data.

During evaluation, the adapter updates the latent predictor with the original training objective on collected transitions. Replay and a parameter anchor constrain the update. A frozen baseline performs no updates. A sham baseline shuffles prediction targets and keeps the rest of the update procedure unchanged. Comparing online and sham updates tests the contribution of correct target pairing. All conditions follow the same sequential evaluation protocol.

\section{Experimental Setup}
\label{sec:setup}
The primary study uses LeWM on TwoRoom navigation, PushT manipulation, OGBench-Cube manipulation, and DMC-Reacher control. We use Cube and Reacher as short names below. TwoRoom, PushT, and Cube use expert-data world models. Reacher uses random-policy training data. Checkpoint comparisons vary training progress within an environment. The coverage experiment changes which transitions are available during training. Table~\ref{tab:configuration} gives the primary configuration.

\begin{table}[t]
\centering
\caption{Primary LeWM configuration. The same cost-stall settings are used across the four tasks. Dose-response and transfer experiments have their own stated seed counts and budgets. The normalized-weight grid uses one seed per setting.}
\label{tab:configuration}
\small
\begin{tabular}{ll@{\qquad}ll}
\toprule
Quantity & Setting & Quantity & Setting \\
\midrule
Model parameters & 18M & Image resolution & $224\times224$ \\
Latent dimension & 192 & CEM samples / elites & 300 / 30 \\
Horizon / action block & 5 / 5 & Refinement cap & 30 \\
Stall threshold & $10^{-2}$ & Patience / minimum depth & 2 / 3 \\
EMA coefficient & 0.5 & Episodes per seed & 50 \\
Initial controller seeds & 3 to 6 & Dose-response seeds per cell & 100 \\
\bottomrule
\end{tabular}
\end{table}

Fixed-depth CEM measures success as refinement increases. The primary stopping rule and its alternatives share the planner configuration. We select the fixed reference from the evaluated depth grid, which means it is not independently tuned. The disagreement experiment varies $\beta$ with the model and rollout procedure fixed. TD-MPC2 \citep{hansen2024tdmpc2} and DINO-WM \citep{zhou2024dinowm} test stopping with a common threshold within each family. TD-MPC2 samples actions using learned latent values. DINO-WM plans with dynamics on fixed visual features.

Evaluations sample start and goal pairs from the available datasets. These samples may overlap with model-training data. Matched seed comparisons reuse start and goal sets, although predictor-dropout randomness is not fully paired. We report mean performance and the seed counts for each comparison. Controller comparisons with unequal seed sets are descriptive. Appendix~\ref{app:protocol} gives the sampling and pairing procedures.

Success measures goal completion on LeWM and DINO-WM tasks. TD-MPC2 uses return relative to each task's best fixed-depth reference. Mean depth measures the refinement iterations used. With candidate count, horizon, and stochastic rollout count fixed, imagined transition evaluations decrease in proportion to depth. Iteration savings do not measure end-to-end runtime. Ensemble-size comparisons change the work per iteration, so we report them separately from the stopping comparison.

For adaptation, we report the fraction of the coverage deficit recovered. The deficit is the frozen-success gap between the volume-matched reference and the hole model. We divide the online improvement by that gap under the same region-goal evaluation. The success-matched checkpoint tests the effect of initial performance and is not the denominator. The recovered fraction can exceed one or become negative because it is measured relative to the chosen reference.

\section{Results}
\subsection{Stopping Performance}
\label{sec:controller}
Cost-stall rules reduce refinement on tasks where additional depth has little effect on success. Table~\ref{tab:controller_main} compares the primary rule with the best fixed-depth reference and best-elite stopping. On TwoRoom, PushT, and Reacher, the primary rule uses less depth than the cap and reaches mean success close to the fixed references. Many solves improve rapidly before stalling. Cases that continue to improve run toward the budget limit.

\begin{table}[t]
\centering
\caption{Controller success (\%, mean across seeds) and mean depth. The primary rule saves 6 to 27\% of iterations on TwoRoom, PushT, and Reacher relative to the cap. Each seed contains 50 episodes. Seed counts for fixed, primary, and best-elite are $(3,5,3)$ on TwoRoom, $(5,5,3)$ on PushT, and $(3,3,3)$ on Cube and Reacher. Fixed depth is selected from the evaluated grid. The seed sets differ between methods.}
\label{tab:controller_main}
\small
\begin{tabular}{lccc}
\toprule
Environment & Best fixed & Primary & Best-elite \\
 & success @ depth & success @ depth & success @ depth \\
\midrule
TwoRoom & $90.0$ @ $30$ & $88.0$ @ $22.0$ & $89.3$ @ $9.3$ \\
PushT & $88.4$ @ $30$ & $87.2$ @ $28.3$ & $87.3$ @ $15.3$ \\
Cube & $67.3$ @ $10$ & $63.3$ @ $28.1$ & $63.3$ @ $15.8$ \\
Reacher & $72.0$ @ $30$ & $74.7$ @ $22.3$ & $74.0$ @ $9.3$ \\
\bottomrule
\end{tabular}
\end{table}

Best-elite cost stall often uses fewer iterations than the primary elite-mean rule, with similar mean success. Removing smoothing or relaxing the threshold also reduces depth. Variance stall is unstable on PushT, and elite-set stability changes abruptly with its threshold. Mean-cost descent is therefore one useful stopping statistic among several. Appendix~\ref{app:stopping} gives the complete comparisons and sensitivity curves.

Instances whose cost stalls quickly stop early. Continued cost improvement uses more of the budget. On PushT, higher initial elite cost is associated with greater allocated depth in the displayed traces. A shared rule can therefore allocate different amounts of computation without a difficulty classifier. This allocation still depends on the model cost, including when further cost reduction harms task success.

\subsection{Disagreement Weights}
\label{sec:beta}
\label{sec:reversed}
Rewarding disagreement lowers frozen-model success at the larger tested weights in every environment. Table~\ref{tab:signals_main} summarizes the reductions and selected penalty gains. A disagreement reward changes the actions selected, but new observations cannot improve a predictor whose weights remain fixed. This may explain the loss in these short-horizon tasks.

\begin{table}[t]
\centering
\caption{Success changes from the $\beta=0$ baseline. Each setting uses 100 matched seeds and 50 episodes per seed with $K=4$ stochastic rollouts. Differences are percentage points (pp). The reward rows give the range across tasks, and the penalty rows show selected positive and negative changes. Table~\ref{tab:beta_audited} gives all weight comparisons.}
\label{tab:signals_main}
\small
\begin{tabular}{llcl}
\toprule
Setting & Environment & $\beta$ & Success change (pp) \\
\midrule
Reward & All four & $+10$ & $-3.5$ to $-54.7$ \\
Reward & All four & $+50$ & $-13.0$ to $-78.1$ \\
\midrule
Penalty & Reacher & $-20,-50$ & $+2.38,+3.34$ \\
Penalty & PushT & $-3$ & $+1.42$ \\
Penalty & Cube & $-20,-10$ & $+1.44,+1.88$ \\
Strong penalty & TwoRoom & $-20,-50$ & $-4.7,-13.4$ \\
Strong penalty & PushT & $-50$ & $-4.5$ \\
\bottomrule
\end{tabular}
\end{table}

Penalizing disagreement gives small gains at selected weights on Reacher, PushT, and Cube. Stronger penalties harm TwoRoom and PushT. The effect depends on both the weight and the task. This cross-task comparison changes the environment and training data together, so it cannot isolate training coverage as the cause.

\label{sec:ladder}
Checkpoint comparisons test whether weaker prediction alone explains the penalty benefit. Large changes in one-step prediction error on TwoRoom do not yield a consistent penalty advantage. Reacher shows gains at several checkpoints and reversals at others. The disagreement-aware stopping variant also changes across checkpoints. Scalar prediction error alone does not explain these patterns. Training time and model state remain coupled in this comparison. Appendix~\ref{app:ladder_diagnostic} gives the checkpoint results.

\subsection{Learning Missing Dynamics}
\label{sec:online}
Online predictor updates recover part of the deficit caused by a coverage hole. Table~\ref{tab:adaptation_main} compares frozen, online, and shuffled-target updates for goals inside the removed region. Online learning improves both hole checkpoints and outperforms shuffled-target updates. This advantage suggests that correct target pairing contributes to the gain. It does not recur in the success-matched and full-data controls.

\begin{table}[t]
\centering
\caption{Coverage intervention on a second independently trained sample. Success is the mean percentage over 50 episodes per seed. Checkpoints e2, e3, and e5 denote training epochs. The matched control is a random-drop e2 checkpoint matched to hole e3 on frozen success. Deficit recovery uses volume-matched e3/e5 references with success $87.2\%/83.2\%$. The update conditions in each column share a seed set.}
\label{tab:adaptation_main}
\small
\begin{tabular}{lccccc}
\toprule
Update & Hole e3 & Hole e5 & Matched & Full e2 & Released \\
Seeds & 5 & 5 & 5 & 5 & 3 \\
\midrule
Frozen & $37.2$ & $44.0$ & $37.2$ & $70.8$ & $75.3$ \\
Online & $56.0$ & $62.8$ & $34.8$ & $76.4$ & $74.0$ \\
Sham & $46.4$ & $55.6$ & $38.0$ & $80.8$ & n/a \\
Online $-$ sham (pp) & $+9.6$ & $+7.2$ & $-3.2$ & $-4.4$ & n/a \\
Deficit recovered & $37.6\%$ & $48.0\%$ & n/a & n/a & n/a \\
\bottomrule
\end{tabular}
\end{table}

Random removal preserves access to the region and matches training-data volume. The success-matched checkpoint tests whether poor initial performance alone explains the benefit of adaptation. Together, these controls support learning dynamics needed by the evaluated goals. Shuffled-target updates are an imperfect control for target pairing because shuffling can also change gradient magnitudes and preserve information about visited states.

Lower prediction loss on collected transitions does not consistently improve task success. Reacher adaptation lacks a replicated benefit across its broad-coverage checkpoints. The frozen disagreement penalty also fails to repair the TwoRoom coverage hole consistently across training samples. Action selection under an existing model and learning missing dynamics therefore need separate evaluation. Appendix~\ref{app:online} gives the adaptation protocol and replications.

\subsection{Model Exploitation}
\label{sec:blindspot}
\label{sec:wall}
Cost-stall rules can continue refining after task success has peaked. Cube reaches its best fixed-depth result before the cap, but the primary controller continues and has lower success. On DINO-WM Wall, success also peaks before the full budget as predicted cost continues to fall. Cost-only halting runs close to the cap and misses the earlier peak (Table~\ref{tab:failure_main}).

\begin{table}[t]
\centering
\caption{DINO-WM Wall with a cap of 30 iterations. Success is the mean over five seeds with 50 starts per seed. The peak is selected from the common-depth mean curve. The fixed-depth $K=5$ setting uses more computation per iteration.}
\label{tab:failure_main}
\small
\begin{tabular}{lcc}
\toprule
Planner setting & Success & Mean depth \\
\midrule
Fixed $K=1$, curve peak & $0.576$ & 11 \\
Fixed $K=1$, full budget & $0.356$ & 30 \\
Cost-only halt, $K=1$ & $0.364$ & 29.04 \\
Fixed $K=5$ & $0.508$ & 30 \\
\bottomrule
\end{tabular}
\end{table}

The tested disagreement guards do not reliably distinguish harmful optimization from productive refinement. Tighter guards sometimes improve Cube but stop prematurely on plateau-shaped tasks. Appendix~\ref{app:guard} reports the guard and detector comparisons. A fixed ensemble improves Wall success by changing candidate scoring and computation, without choosing an earlier stopping point.

Extra iterations on a plateau waste computation. After a success peak, they can also worsen the chosen action. Appendix~\ref{app:regret} formalizes the difference using an idealized success curve and constructs identical cost traces with incompatible optimal depths. A stopping guarantee needs a link between internal progress and external success.

\subsection{Transfer Across Planners}
\label{sec:crossfamily}
The stall statistic also reduces refinement outside the primary CEM planner. TD-MPC2 retains most of the best-grid return across the evaluated DMControl tasks at a small mean depth. DINO-WM PushT reaches the same observed mean success as its fixed-cap reference with fewer iterations (Table~\ref{tab:transfer_main}). The threshold is shared within each family. Threshold selection was not evaluated on a separate held-out set.

\begin{table}[t]
\centering
\caption{Cross-family stopping results. TD-MPC2 averages tasks equally and normalizes return by each task's best fixed-depth mean on the same evaluation grid. DINO-WM uses five seeds and cap 30. Detailed task results and configurations appear in Tables~\ref{app:tdmpc2} and~\ref{app:dino}.}
\label{tab:transfer_main}
\small
\begin{tabular}{lll}
\toprule
Family and task & Performance & Mean depth \\
\midrule
TD-MPC2, 19 tasks & $98.9\%$ mean retained return & $3.8$ \\
TD-MPC2, lowest retention & $96.1\%$ retained return & n/a \\
DINO-WM PushT, fixed & $0.852$ success & $30$ \\
DINO-WM PushT, halt & $0.852$ success & $26.44$ \\
\bottomrule
\end{tabular}
\end{table}

DINO-WM contains both PushT, where stopping saves computation, and Wall, where further optimization exploits model errors. Planner identity alone does not predict whether cost-stall stopping will help. The comparison requires measuring task success across depths as well as the computation saved.

\subsection{Ablation Studies}
Stopping earlier reduces refinement with the candidate objective unchanged. Increasing the stochastic rollout count changes both candidate scoring and computation per iteration. On the evaluated LeWM tasks, averaging multiple imagined futures is not necessary for the observed plateau performance. On Wall, a fixed ensemble changes success substantially. Depth comparisons therefore need to account for the rollout count.

Goal distance and terminal prediction variance have different magnitudes. We normalize these magnitudes within the candidate batch before applying a dimensionless weight. Larger positive weights still harm performance. This result suggests that the reward failure is not confined to the original raw weighting. The normalized grid uses fewer seeds than the uniform weight experiment, so the result remains exploratory. Appendix~\ref{app:scale} gives the normalization rule, seed count, and complete values.

For a value-based planner, a large common offset can dominate predicted value magnitudes without changing candidate rankings. Normalizing disagreement by that magnitude can overwhelm the differences used to select actions. Candidate-value spread measures the ranking scale more directly. The partial TD-MPC2 weight grid shows this problem. Its shared-threshold stopping experiment instead measures relative progress in the original objective.

Under the same coverage-hole condition, we compare fixed and adaptive allocation with frozen and online predictors. Online updates improve success under both allocation rules, and adaptive stopping lowers success with either predictor. The mean interaction is small, but these evaluations do not establish independent effects of stopping and learning. Table~\ref{tab:regret} gives the factorial results.

\subsection{Success Across Depths}
A stopping rule observes optimization progress to decide how long to refine an action. Its purpose is to preserve task performance. Consider an idealized episode that uses a single depth at every control step. Let $s(d,x)$ be its success probability at depth $d$ for instance $x$, and let $d_p(x)$ be the earliest depth with maximal success. The relevant error is the difference between success at that optimum and success at the selected depth. Final planning cost measures a different quantity.

Suppose success is nondecreasing before $d_p(x)$ and nonincreasing afterward. Early stopping loses progress available before the peak. Late stopping loses any success that declines after it. On a plateau, the latter loss is zero even when the planner spends unnecessary iterations. On a declining curve, cost can keep improving as success falls. A conservative stopping rule can therefore save computation on a plateau and still fail to protect performance after a peak.

Appendix~\ref{app:regret} proves this separation and constructs instances with identical cost traces but different optimal depths. A rule that reads only those traces must make the same decision on both instances, so at least one incurs a success loss. The statement concerns an idealized curve with a single depth throughout an episode. Actual receding-horizon execution selects depth repeatedly. Stopping in that setting may require observations that connect model predictions with their realized consequences.

\section{Conclusion}
Cost stalling reduces refinement when success plateaus with depth. Disagreement penalties help at selected task and weight settings, and online predictor updates recover part of a deficit caused by missing training coverage. These gains come from different changes to planning. When continued optimization exploits model errors, falling cost no longer indicates a better action. Stopping safely after a success peak requires information beyond the cost trajectory.

\clearpage
\bibliographystyle{iclr2027_conference}
\bibliography{references}

\clearpage
\appendix
\section{Evaluation Details}
\label{app:protocol}
\subsection{Pairing and Compute}
Each paired comparison reuses the starts and goals sampled by an evaluation seed. If $S_{r,A}$ and $S_{r,B}$ are success rates for seed $r$ under configurations $A$ and $B$, the mean difference is
\begin{equation}
 \widehat\Delta_{A-B}=\frac{1}{R}\sum_{r=1}^{R}(S_{r,A}-S_{r,B}).
 \label{eq:paired_main}
\end{equation}
Each experiment retains its seed count and evaluation protocol. Repeating episodes under a fixed seed does not reproduce variation across independently sampled seeds. Repeated evaluations also leave training variation unchanged.

Depth has a similarly specific interpretation. If each refinement iteration evaluates $S$ action candidates with $K$ stochastic rollouts over horizon $H$, a simple count of imagined transition evaluations is proportional to
\begin{equation}
 B_{\mathrm{imag}}(x)=S K H\,\hat n(x),\qquad
 \eta_{\mathrm{iter}}=1-\frac{\mathbb E_x[\hat n(x)]}{N}.
 \label{eq:compute_main}
\end{equation}
The second expression is the iteration saving relative to a fixed cap with the same per-iteration configuration. It is not an end-to-end runtime measurement. Encoding, action execution, batched stopping, and any uncertainty or online-update overhead contribute additional cost. Increasing the rollout count can consume more computation even if the chosen depth falls. Our principal compute comparison therefore reports used iterations alongside success and keeps ensemble-size changes in the ablations.

\subsection{Regret and Coverage}
\label{sec:regret}
The following success curves describe the success-rate tasks. Let $x$ denote a sampled start and goal pair, and let $\omega$ denote planner randomness. Fixed-depth episode curves use the same depth at every control step. In the idealization, $\tilde d(x,\omega)\in\{n_{\min},\ldots,N\}$ selects one depth for the episode. Let $s(d,x)\in[0,1]$ denote success probability at integer depth $d$. Define $d^*(x)\in\arg\max_{d\in\{n_{\min},\ldots,N\}}s(d,x)$ as an allowed depth that maximizes success. The adaptive stopping regret is
\begin{equation}
 r_{\mathrm{stop}}(x,\omega)=s(d^*(x),x)-s(\tilde d(x,\omega),x),\quad
 \mathcal R_{\mathrm{stop}}=\mathbb E_{x,\omega}[r_{\mathrm{stop}}(x,\omega)].
 \label{eq:stopmain}
\end{equation}
This uniform-depth oracle does not measure our controller's sequence of per-step decisions. Identifying curve-adaptive success with observed policy success requires conditional independence of selected depth and outcomes given the instance. The common-fixed-depth population gap lower-bounds the curve-based oracle gap only when both use the same adaptive term. Selecting the fixed-depth reference after evaluation can bias the measured gap upward. TD-MPC2 returns use a different scale and are excluded from this formalism.

A separate descriptive gap concerns training coverage (Section~\ref{sec:online}, including the distinct matched controls). Let $S_{\mathrm{match}}$ be success of the volume-matched frozen reference, $S_{\mathrm{hole}}$ success of the hole model, and $S_{\mathrm{online}}$ success after adaptation under the same region-goal protocol. Then
\begin{equation}
 G_{\mathrm{hole}}=S_{\mathrm{match}}-S_{\mathrm{hole}},\qquad
 \rho_{\mathrm{recovered}}=\frac{S_{\mathrm{online}}-S_{\mathrm{hole}}}{G_{\mathrm{hole}}}.
 \label{eq:coverage_main}
\end{equation}
The ratio depends on the adaptation budget and the chosen reference. It may be negative or exceed unity, and becomes unstable when the reference gap is close to zero. It describes the recovered success gap without bounding other sources of model error. The factorial and coverage comparisons use separate online evaluations under the same condition label. We retain the seed pairing within each comparison.

Appendix~\ref{app:regret} constructs identical observable cost traces with incompatible optimal depths, giving positive worst-case regret for deterministic trace-only rules and a separate randomized bound. Exact equality is a construction assumption, not established for Cube or Wall. The tested detectors also have limited success. Their results leave open whether other stopping signals can identify model exploitation.

\subsection{Post-peak Loss}
The success curve supplies a precise distinction between harmless extra computation and damaging over-optimization. Under the appendix's weakly unimodal shape assumption (nondecreasing before the earliest peak and nonincreasing afterward), let $d_p(x)$ be the earliest maximizing depth. Define the pre-peak shortfall and post-peak decline by
\begin{equation}
 G(x)=s(d_p(x),x)-s(n_{\min},x),\qquad
 D(x)=s(d_p(x),x)-s(N,x).
 \label{eq:shape_main}
\end{equation}
Both are nonnegative, but they describe different errors. An early stop can forfeit progress before the optimum. A late stop can hurt only to the extent that success declines afterward. Monotonicity on each side of the earliest peak gives the pointwise inequality
\begin{equation}
 r_{\mathrm{stop}}(x,\omega)\le
 G(x)\mathbf1[\tilde d(x,\omega)<d_p(x)]
 +D(x)\mathbf1[\tilde d(x,\omega)>d_p(x)].
 \label{eq:shape_bound_main}
\end{equation}
Taking expectations gives Appendix~\ref{app:regret}'s early/late bound. A conservative rule can waste iterations on a plateau yet preserve success. The same delay can incur loss on a declining curve. This explanation assumes a success curve for each instance. The measured aggregate curves do not determine the episode regret of a controller that changes depth at every step.

\subsection{Scope and limitations}
\label{sec:limits}
The normalized-weight and detector comparisons include small-seed and incomplete grids. They describe the tested configurations. They do not establish a calibrated uncertainty measure or replicate every contrast across planner families.

The experiments use near-fully-observed tasks and short receding-horizon control with a moderate-scale world model. In partially observed tasks, information can improve the belief state without updating predictor parameters. That benefit remains untested here. Online adaptation uses a particular update procedure and coverage manipulation. Behavioral evaluation samples may overlap with training data. The sham control preserves anchor supervision but can change gradient norms. We share thresholds within each planner family without a separate held-out selection protocol. These choices limit transfer to other data distributions, update rules, and planning horizons.

\section{Supplementary Results}
\label{app:supplementary_measurements}
These figures supplement the analyses of compute allocation, disagreement weighting, and over-optimization. Captions specify the evaluation protocol for each experiment.

\begin{figure}[t]
\centering
\includegraphics[width=\linewidth]{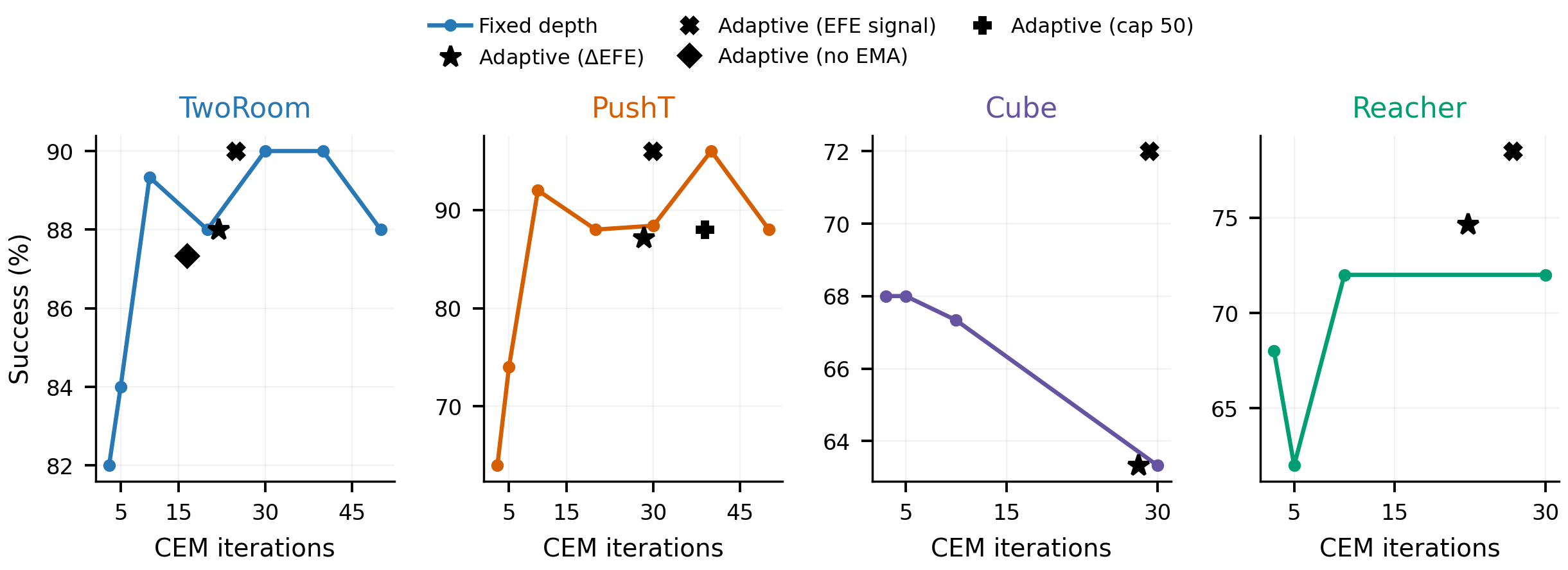}
\caption{Fixed-depth success and adaptive halting plotted against mean refinement depth. Figure~\ref{fig:pipeline} shows the planning procedure.}
\label{fig:depth}
\end{figure}

\begin{figure}[!htbp]
\centering
\begin{subfigure}[t]{0.58\linewidth}
\includegraphics[width=0.76\linewidth]{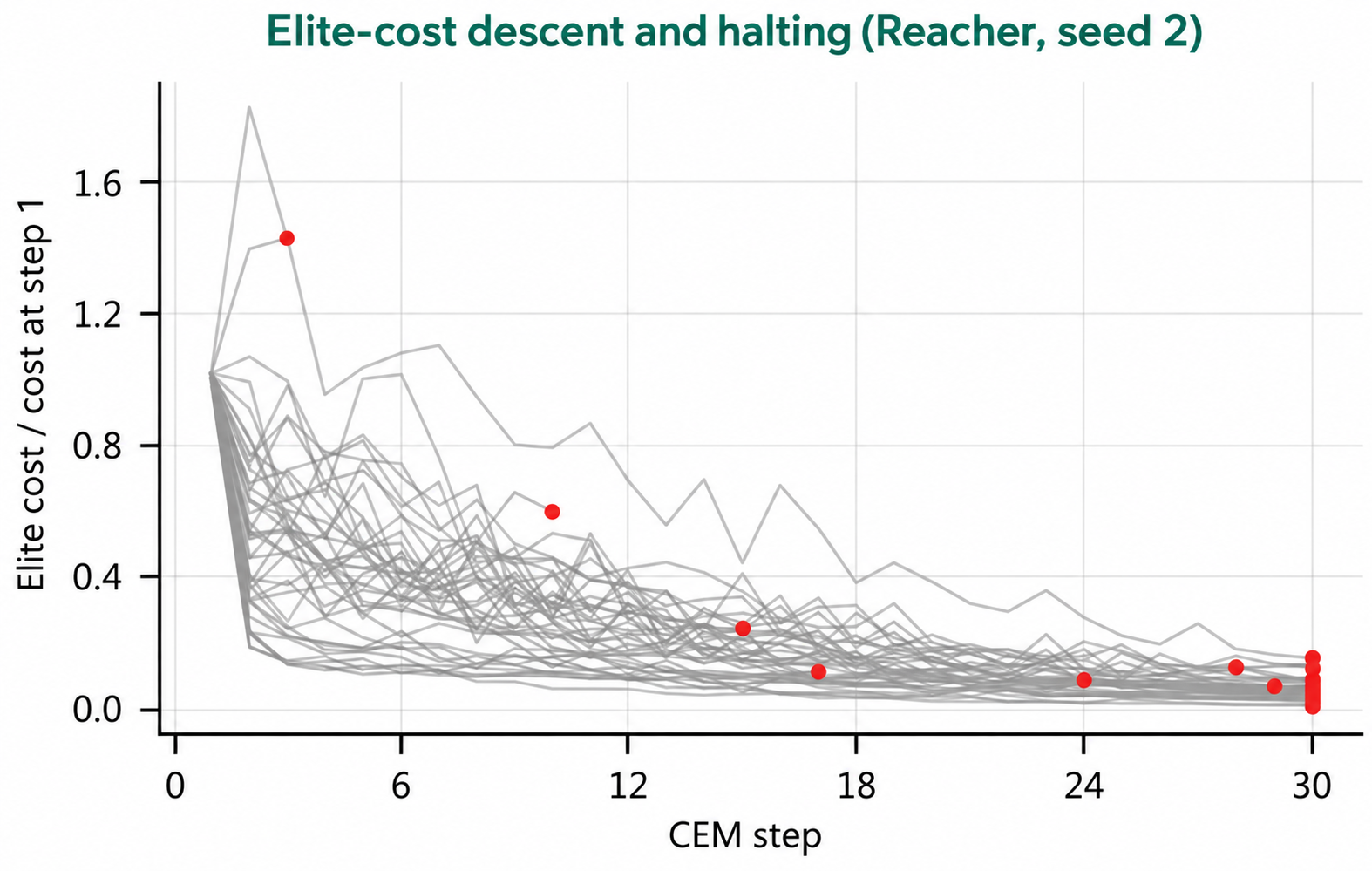}
\end{subfigure}\hfill
\begin{subfigure}[t]{0.39\linewidth}
\includegraphics[width=0.76\linewidth]{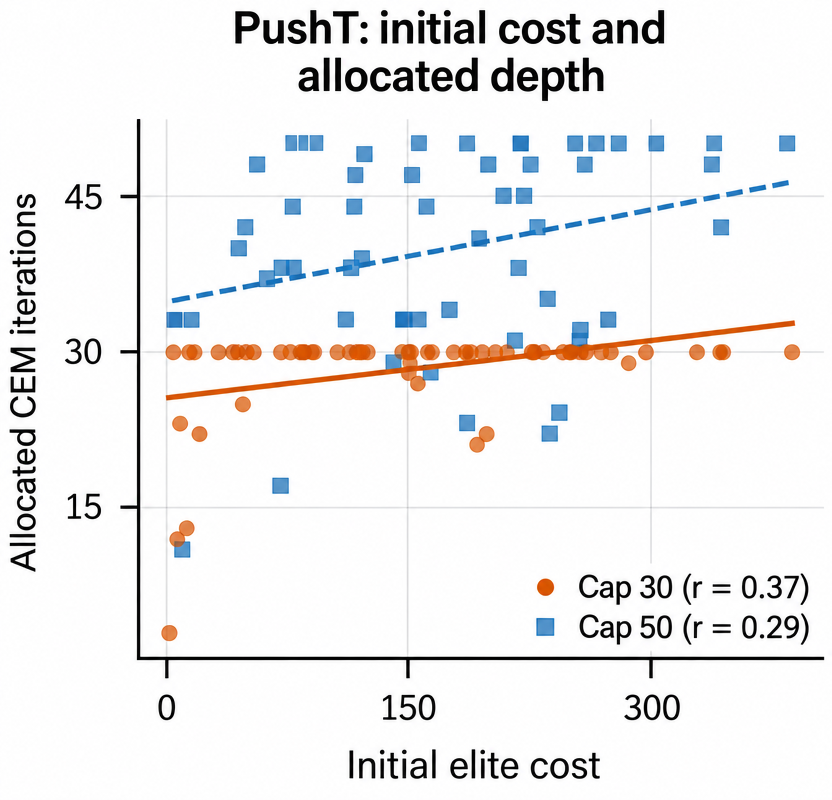}
\end{subfigure}
\caption{Elite-cost trajectories and allocated depth. Left, example trajectories show rapid early descent and a slower-descending case that receives more iterations. Red markers indicate halting. Right, initial elite cost correlates with depth on PushT ($r{\approx}0.3$). Initial cost is a model-based difficulty proxy. The correlation does not use an independently measured difficulty label.}
\label{fig:descent}
\end{figure}

\begin{figure}[!htbp]
\centering
\includegraphics[width=\linewidth]{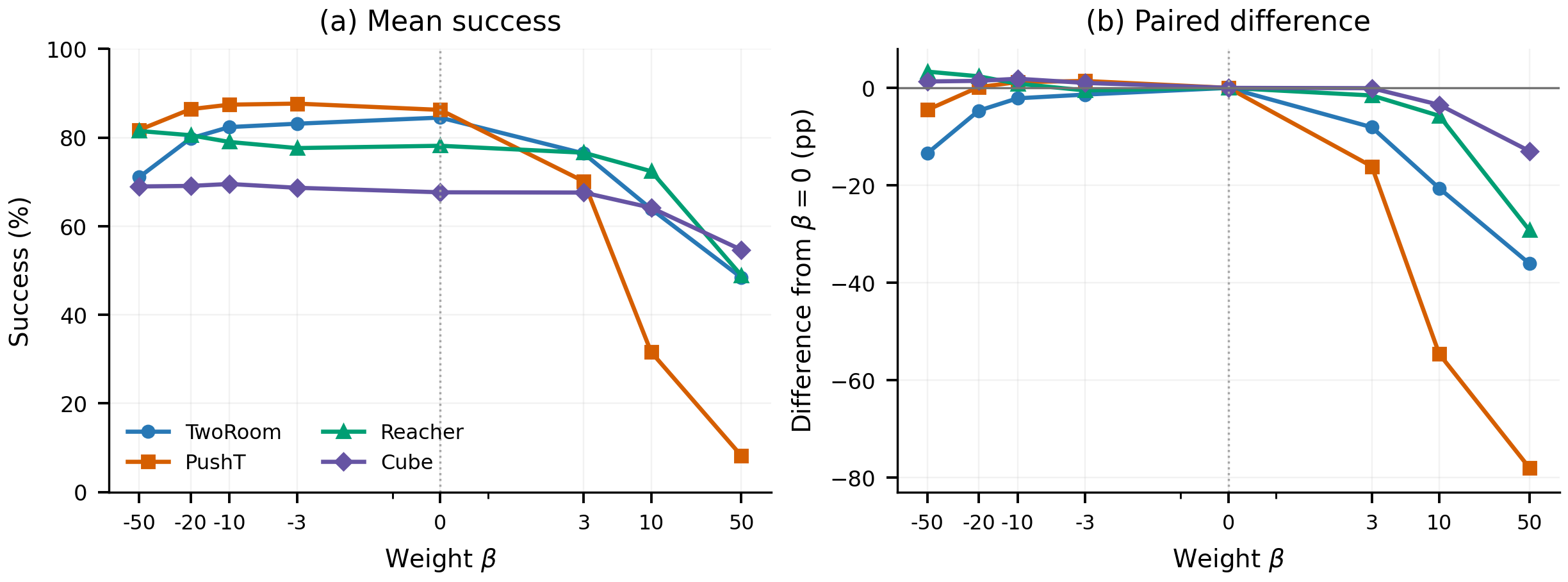}
\caption{Disagreement-weight sweep with 100 seeds per cell and 50 episodes per seed. (a) Mean success against $\beta$ on a symlog axis. Larger positive weights reduce success in every environment. Some negative weights give small gains on Reacher, PushT, and Cube. Strong penalties harm TwoRoom and PushT. (b) Mean paired differences from $\beta{=}0$ (Table~\ref{tab:beta_audited}). The gains depend on the environment and weight. Coverage is not independently varied.}
\label{fig:beta}
\end{figure}

\begin{figure}[!htbp]
\centering
\includegraphics[width=\linewidth]{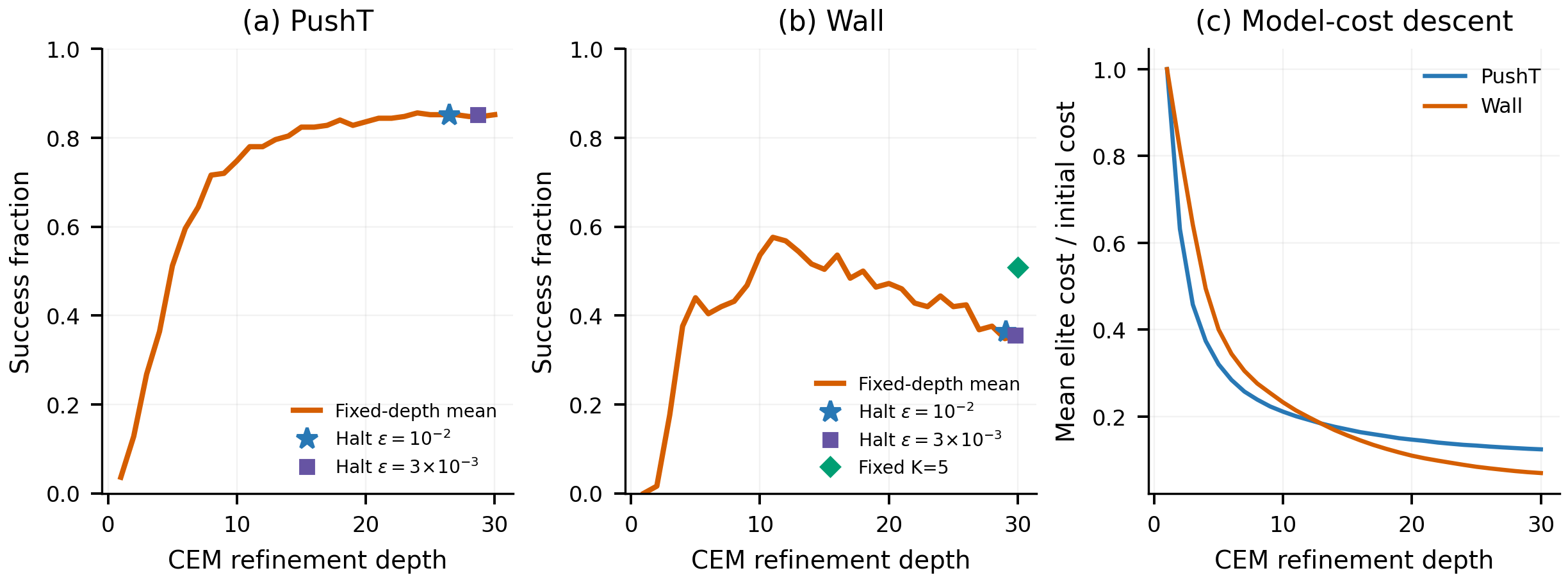}
\caption{DINO-WM depth curves with five seeds and a cap of 30. Wall's mean success peaks at $0.576$ at iteration 11. It falls to $0.356$ at iteration 30, a $38.2\%$ relative decline. The peak is selected from the mean curve at a common depth across seeds. Cost-only halting with $\varepsilon=10^{-2}$ attains $0.364$ success at mean depth $29.04$. The fixed-depth $K=5$ ensemble attains $0.508$. The ensemble uses more stochastic rollouts at a fixed depth.}
\label{fig:wall}
\end{figure}

\section{Checkpoint Comparisons}
\label{app:ladder_diagnostic}

Cross-environment contrasts conflate training data with task differences. We therefore compare checkpoints within a training trajectory for TwoRoom and Reacher. The environment and planner protocol remain fixed. Each checkpoint has a held-out one-step prediction error and evaluations of fixed-depth and adaptive planning (Table~\ref{app:ladder}). Training time and model state vary together. Coverage is not independently manipulated.

The penalty's effect changes with the checkpoint. TwoRoom's fixed-depth success curves stay close as probe error falls, and their difference changes sign. Reacher gains at several checkpoints but loses success at e8. Its largest gain coincides with the e3 increase in probe error. Training time and model state change together, so this association does not identify either probe error or coverage as the cause.

The relative performance of the stopping rules also changes during training. At Reacher e8, epistemic-aware halting exceeds cost-only halting but remains below fixed-depth planning. At e12, cost-only halting has the higher mean. These reversals argue against treating epistemic-aware halting as a consistent improvement.

\begin{figure}[t]
\centering
\includegraphics[width=0.82\linewidth]{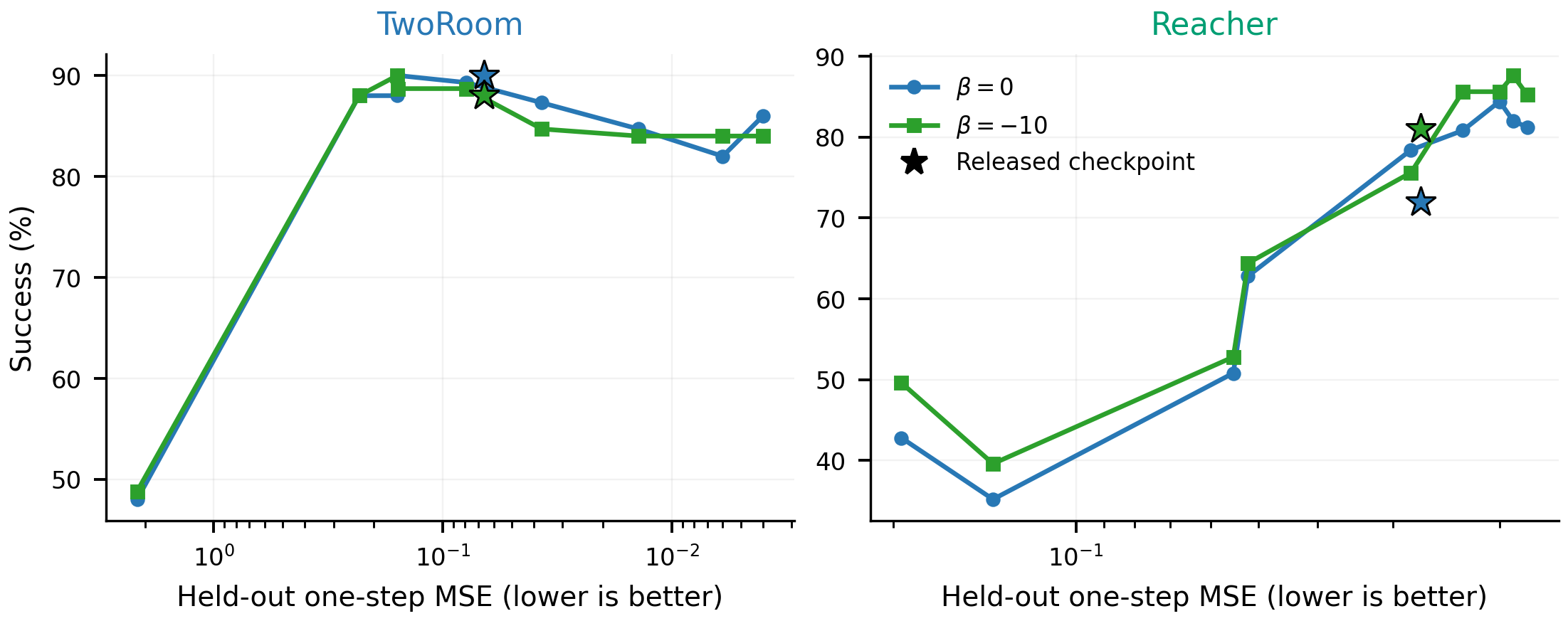}
\caption{Checkpoint quality and uncertainty penalties. Fixed-depth success at $\beta{=}0$ and $\beta{=}{-}10$ is plotted against held-out one-step MSE on a log axis. Means are listed in Table~\ref{app:ladder}. TwoRoom's curves remain close across the measured range. Reacher has positive contrasts at several checkpoints and a negative contrast at e8. Coverage is not independently varied. Stars mark released-model anchors with fewer seeds than the separate $n{=}100$ weight sweep.}
\label{fig:ladder}
\end{figure}

\section{Candidate Selection}
\label{app:conceptual_guides}
The diagram shows how disagreement changes candidate selection and how a separate stopping rule controls refinement.

\begin{figure}[t]
\centering
\includegraphics[width=0.92\linewidth]{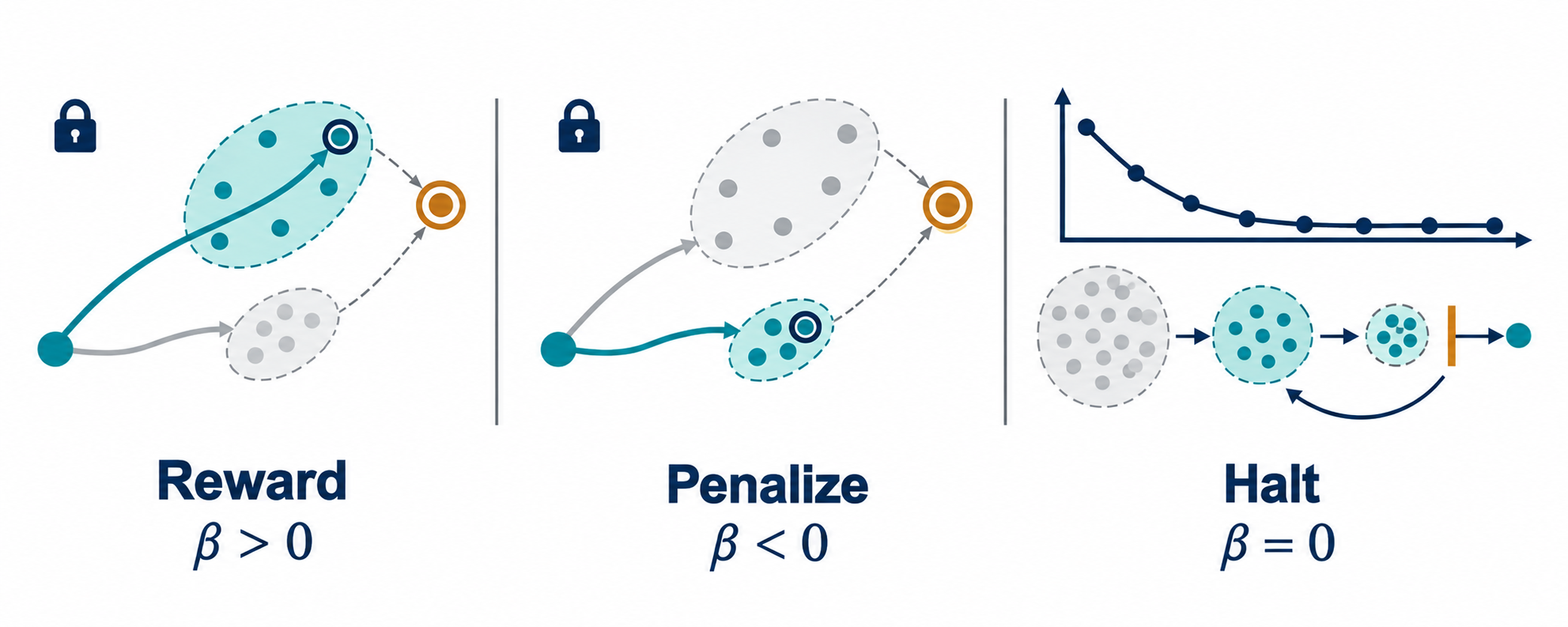}
\caption{Candidate preference and compute allocation. Positive $\beta$ favors terminal rollout spread and negative $\beta$ penalizes it. Highlighted bundles show candidate selection. Penalties can help or hurt task success. Cost-only halting ($\beta=0$) controls refinement depth. Disagreement is an uncalibrated surrogate. Schematic.}
\label{fig:roles}
\end{figure}

\section{Full Results Tables}
\label{app:tables}
The tables report checkpoint, transfer, and adaptation experiments. Each experiment retains its own seed count and evaluation protocol. Trace measurements and checkpoint sweeps use separate evaluations from the disagreement-weight comparison.

\begin{table}[htbp]
\centering
\caption{Within-environment checkpoint comparison. Per-epoch LeWM checkpoints, 50 episodes per cell. Probe MSE, $\beta{=}0$/$\beta{=}{-}10$ success, and adaptive halting. See Fig.~\ref{fig:ladder} for visualization.}
\label{app:ladder}
\small
\setlength{\tabcolsep}{5pt}
\begin{tabular}{llccccc}
\toprule
Env & Ckpt & Probe MSE & $\beta{=}0$ & $\beta{=}{-}10$ & adaCost & adaEFE \\
\midrule
TwoRoom & e1  & $2.161$ & $48.0$ & $48.8$ & $45.3$ & $42.7$ \\
        & e2  & $0.232$ & $88.0$ & $88.0$ & $87.3$ & $89.3$ \\
        & e3  & $0.157$ & $90.0$ & $88.7$ & $90.7$ & $92.0$ \\
        & e5  & $0.158$ & $88.0$ & $90.0$ & $90.0$ & $90.7$ \\
        & e8  & $0.079$ & $89.3$ & $88.7$ & $89.3$ & $89.3$ \\
        & e12 & $0.037$ & $87.3$ & $84.7$ & $85.3$ & $87.3$ \\
        & e18 & $0.014$ & $84.7$ & $84.0$ & $84.7$ & $86.0$ \\
        & e25 & $0.006$ & $82.0$ & $84.0$ & $86.0$ & $86.7$ \\
        & e30 & $0.004$ & $86.0$ & $84.0$ & $86.0$ & $87.3$ \\
\midrule
Reacher & e1  & $0.137$ & $35.2$ & $39.6$ & $44.0$ & $38.0$ \\
        & e2  & $0.055$ & $50.8$ & $52.8$ & $46.7$ & $44.0$ \\
        & e3  & $0.194$ & $42.8$ & $49.6$ & $42.0$ & $44.0$ \\
        & e5  & $0.052$ & $62.8$ & $64.4$ & $62.0$ & $60.7$ \\
        & e8  & $0.028$ & $78.4$ & $75.6$ & $70.7$ & $76.0$ \\
        & e12 & $0.023$ & $80.8$ & $85.6$ & $85.3$ & $78.7$ \\
        & e18 & $0.020$ & $84.4$ & $85.6$ & $80.7$ & $79.3$ \\
        & e25 & $0.019$ & $82.0$ & $87.6$ & $82.7$ & $80.0$ \\
        & e30 & $0.018$ & $81.2$ & $85.2$ & $82.7$ & $81.3$ \\
\bottomrule
\end{tabular}
\end{table}

\begin{table}[htbp]
\centering
\caption{TD-MPC2 evaluation on nineteen DMControl tasks. Each task uses one released checkpoint and four seeds $\{2,3,4,5\}$, with 30 episodes per seed. Returns are means across seeds. The fixed reference maximizes mean return over $N\in\{3,6,9,12\}$ on the evaluation seeds. Halting uses $\varepsilon=3\times10^{-3}$. Its cap is 8 on dog-walk, humanoid-run, and humanoid-walk, and 6 on the remaining tasks. Relative return is $100\times\text{halt mean}/\text{fixed mean}$, so values above $100\%$ indicate greater return than the fixed reference. Depth is averaged over seeds and summary rows weight tasks equally. Calibration evaluations are excluded. The fixed reference is selected retrospectively.}
\label{app:tdmpc2}
\small
\setlength{\tabcolsep}{3pt}
\begin{tabular}{lrcccc}
\toprule
Task & $N^*$ & Fixed return & Halt return & Relative return & Depth \\
\midrule
acrobot-swingup & 6 & $661.2$ & $661.8$ & $100.1\%$ & $3.1$ \\
cheetah-run & 9 & $861.1$ & $828.1$ & $96.2\%$ & $3.5$ \\
cup-catch & 9 & $981.1$ & $979.1$ & $99.8\%$ & $3.0$ \\
dog-walk & 9 & $799.2$ & $783.8$ & $98.1\%$ & $5.3$ \\
finger-spin & 6 & $988.3$ & $985.7$ & $99.7\%$ & $3.8$ \\
finger-turn-hard & 3 & $972.7$ & $967.0$ & $99.4\%$ & $3.0$ \\
fish-swim & 12 & $787.5$ & $775.7$ & $98.5\%$ & $3.2$ \\
hopper-hop & 12 & $570.1$ & $568.0$ & $99.6\%$ & $5.9$ \\
hopper-stand & 6 & $961.3$ & $952.9$ & $99.1\%$ & $3.0$ \\
humanoid-run & 12 & $680.9$ & $659.6$ & $96.9\%$ & $7.4$ \\
humanoid-walk & 12 & $926.5$ & $890.4$ & $96.1\%$ & $5.3$ \\
pendulum-swingup & 12 & $846.1$ & $846.0$ & $100.0\%$ & $3.0$ \\
quadruped-run & 6 & $957.8$ & $953.3$ & $99.5\%$ & $3.7$ \\
quadruped-walk & 9 & $979.8$ & $976.0$ & $99.6\%$ & $3.1$ \\
reacher-easy & 3 & $977.9$ & $978.7$ & $100.1\%$ & $3.0$ \\
reacher-hard & 6 & $981.9$ & $981.8$ & $100.0\%$ & $3.0$ \\
walker-run & 9 & $891.2$ & $867.9$ & $97.4\%$ & $3.6$ \\
walker-stand & 6 & $994.8$ & $994.3$ & $99.9\%$ & $3.0$ \\
walker-walk & 9 & $985.3$ & $984.1$ & $99.9\%$ & $3.0$ \\
\midrule
\multicolumn{4}{l}{Mean over 19 tasks} & $98.9\%$ & $3.8$ \\
\multicolumn{4}{l}{Worst retained return} & $96.1\%$ & $\cdot$ \\
\bottomrule
\end{tabular}
\end{table}

\begin{table}[htbp]
\centering
\caption{DINO-WM evaluation with five seeds (1 to 5), 50 starts per seed, and cap 30. Success is the mean across seeds. Depth counts the iterations used by each trajectory. The $K=5$ configuration uses fixed-depth ensemble scoring.}
\label{app:dino}
\small
\begin{tabular}{lcc}
\toprule
Configuration & PushT success @ depth & Wall success @ depth \\
\midrule
Fixed $K=1$ & $0.852$ @ $30$ & $0.356$ @ $30$ \\
Halt $\varepsilon=10^{-2}$ & $0.852$ @ $26.44$ & $0.364$ @ $29.04$ \\
Halt $\varepsilon=3\times10^{-3}$ & $0.852$ @ $28.71$ & $0.356$ @ $29.79$ \\
Fixed $K=5$ & $\cdot$ & $0.508$ @ $30$ \\
\bottomrule
\end{tabular}
\end{table}

\section{Regret Analysis}
\label{app:regret}

\subsection{Formal setup}

The analysis uses the uniform-depth episode idealization of Appendix~\ref{sec:regret}. A single depth is applied at every control step. The evaluated controller instead chooses a sequence of depths. The bounds therefore describe success curves rather than its episode-policy regret.

All depths are integer-valued. Fix a refinement budget $N$ and minimum depth $n_{\min}$.
A fixed-depth policy uses $d$ iterations for every instance.
An adaptive rule chooses $\hat d(x,\omega)$ for instance $x$ and planner randomness $\omega$,
with $n_{\min}\le\hat d\le N$. We suppress $\omega$ in the notation below.
Expectations and probabilities include this randomness unless the rule is deterministic.
Let $s(d,x)\in[0,1]$ denote success probability at fixed depth $d$.
Let $c_t(x)$ denote the elite planning cost at iteration $t$.
The controller observes cost trajectories but cannot observe $s$ at deployment.

Per-instance stopping regret and its average over a test distribution
$\mathcal{D}$ are
\[
  r_{\mathrm{stop}}(x) = s(d^*(x), x) - s(\hat{d}(x), x),
  \qquad
  \mathcal{R}_{\mathrm{stop}} = \mathbb{E}_{x \sim \mathcal{D},\omega}\!\left[ r_{\mathrm{stop}}(x) \right],
\]
where $d^*(x)$ is the smallest maximizer of $s(d,x)$ over allowed integer depths.

\paragraph{Relation to adaptive execution.} The selected depth can depend on the solver's random trajectory.
Under conditional independence of selected depth and potential outcomes given $x$,
the adaptive success probability equals $\mathbb E_{x,\omega}[s(\hat d(x),x)]$.
Without that condition, selecting depth can also select outcomes.
This effect is distinct from allocating different depths to different instances.
Instance-dependent allocation alone can outperform a common fixed depth.
The curve-based regret is nonnegative by definition.
The $\Delta$ column in Table~\ref{tab:regret} instead compares marginal means with a common fixed-depth reference.
That descriptive difference can be negative.

\paragraph{Unimodal depth structure.} Say $s(\cdot,x)$ is weakly unimodal
with onset
$d_p(x) = \min\{d \in [n_{\min},N] : s(d,x) = \max_{d'\in[n_{\min},N]\cap\mathbb{Z}} s(d',x)\}$
if it is nondecreasing on $[n_{\min}, d_p(x)]$ and nonincreasing on
$[d_p(x), N]$. Selecting the earliest maximizer also defines early and late stopping when the optimum is flat.
Then $d^*(x) = d_p(x)$. Define the
pre-plateau shortfall and post-plateau decline
\[
  G(x) = s(d_p(x),x) - s(n_{\min},x) \ge 0,
  \qquad
  D(x) = s(d_p(x),x) - s(N,x) \ge 0 .
\]
$D(x)=0$ for a flat post-peak plateau. A positive $D(x)$ permits success loss from late stopping.

\paragraph{Proposition 1.} Stopping regret under unimodal structure. Let
$q_e = \Pr_{x,\omega}[\hat{d}(x) < d_p(x)]$ and $q_\ell = \Pr_{x,\omega}[\hat{d}(x) > d_p(x)]$
be the early- and late-halt probabilities of an adaptive rule. If $s(\cdot,x)$
is weakly unimodal for every $x$, then
\begin{equation}
  \mathcal{R}_{\mathrm{stop}}
  \;\le\; q_e \cdot \mathbb{E}\!\left[ G(x) \,\middle|\, \hat{d} < d_p \right]
  \;+\; q_\ell \cdot \mathbb{E}\!\left[ D(x) \,\middle|\, \hat{d} > d_p \right].
  \label{eq:prop1}
\end{equation}

Proof. Condition on the halt event. If $\hat{d}(x) = d_p(x)$ the regret
is zero. If $\hat{d}(x) < d_p(x)$, monotonicity on $[n_{\min}, d_p]$ gives
$s(\hat{d},x) \ge s(n_{\min},x)$, so
$r_{\mathrm{stop}}(x) \le s(d_p,x) - s(n_{\min},x) = G(x)$. If
$\hat{d}(x) > d_p(x)$, monotonicity on $[d_p, N]$ gives
$s(\hat{d},x) \ge s(N,x)$, so $r_{\mathrm{stop}}(x) \le D(x)$.
Taking expectations and applying the law of total expectation yields
Eq.~\ref{eq:prop1}, with a zero-probability event contributing zero (equivalently use expectations of $G$ and $D$ times their event indicators). $\square$

The bound separates loss from early stopping and post-peak decline.
If a rule never stops early and $D(x)=0$ for every instance, its curve-based regret is zero.
Otherwise the bound depends on the stopping probabilities and conditional shortfalls.
These quantities depend on the instance distribution.

We do not estimate $q_e$ or $q_\ell$, which require per-instance oracle depths.
The measured grids use a small set of fixed depths, with some cells evaluated at a single seed.
They show aggregate plateaus or declines but do not establish unimodality for every instance.
In particular, a best-fixed-minus-adaptive success gap does not estimate $D(x)$.
The Cube and Wall curves illustrate declining aggregate success after a peak.
They do not supply the per-instance quantities needed to evaluate the bound.

\subsection{Cost-based Stopping Limit}

Proposition 1 bounds regret in terms of $q_e, q_\ell$, which depend on the
rule's ability to locate $d_p(x)$. But the rule sees only the cost trajectory.
Call a halting rule cost-internal if $\hat{d}(x)$ is a measurable
function of $\{c_t(x)\}_{t \le N}$ alone, where $c_t$ is the pragmatic
elite cost minimized by the primary controller at $\beta{=}0$.
Our $\deltaefe$ rule and its patience and threshold variants are cost-internal.
The disagreement guards in Appendix~\ref{app:guard} use additional information.

Write $x\sim x'$ when $c_t(x)=c_t(x')$ for all $t\le N$.
Such instances are indistinguishable to a cost-internal rule.
The following construction gives a worst-case lower bound within this class.

\paragraph{Proposition 2.} Positive worst-case regret for cost-internal halting. Fix $\delta, \delta^{\prime} \in (0,1]$ and choose $v\in[0,1-\max\{\delta,\delta^{\prime}\}]$.
Fix depths $n_{\min} \le a < b \le N$. There
exists a pair $x \sim x'$ of weakly unimodal instances with
\[
  s(d,x) = \begin{cases} v & d < a \\ v{+}\delta & a \le d < b \\ v{+}\delta{-}\delta_0 & d \ge b \end{cases}
  \qquad
  s(d,x') = \begin{cases} v & d < b \\ v{+}\delta' & d \ge b \end{cases}
\]
(any $0 < \delta_0 \le \delta$) such that every deterministic cost-internal rule
satisfies
\begin{equation}
  \max\{\, r_{\mathrm{stop}}(x),\; r_{\mathrm{stop}}(x') \,\} \;\ge\; \min\{\delta_0,\, \delta'\} \;>\; 0 .
  \label{eq:prop2}
\end{equation}

Proof. A cost-internal rule reads only $\{c_t\}$, which is identical on
$x$ and $x'$. It therefore chooses the same depth $\hat d$ for both.
The earliest oracle depths are $d^*(x)=a$ and $d^*(x^{\prime})=b$. If $\hat{d} < b$ then on
$x'$ the rule attains $v$ against an oracle value $v + \delta'$, so
$r_{\mathrm{stop}}(x') \ge \delta'$. If $\hat{d} \ge b$ then on $x$ it attains
$v + \delta - \delta_0$ against oracle $v + \delta$, so
$r_{\mathrm{stop}}(x) \ge \delta_0$. No $\hat{d}$ avoids both. $\square$

Proposition 2 gives a worst-case construction rather than a lower bound for every instance.
Flat success curves can have zero regret at every depth.
The construction shows that cost traces alone cannot guarantee optimal depth without a link between cost and external success.

Exact equality of cost traces is an assumption of the construction.
The Cube and Wall experiments do not establish such an indistinguishable pair.
Similar traces also need not yield similar decisions because threshold rules can be discontinuous.
An approximate version would require a bound on observational distinguishability.
For randomized rules, the corresponding worst-case expected regret is at least
$\delta_0\delta^{\prime}/(\delta_0+\delta^{\prime})$.
This follows by balancing the probabilities of stopping before and after $b$.

Disagreement is an additional observation, so Equation~\ref{eq:prop2} does not cover guards that use it.
The fixed-depth ensemble exceeds cost-only halting in four of five Wall seeds.
That comparison changes candidate scoring and computation, rather than testing disagreement-based stopping.
Appendix~\ref{app:guard} evaluates such stopping signals separately.

\subsection{Reference-based coverage deficit}

Let $\theta_{\mathcal H}$ be a model trained with a spatial coverage hole.
Its coverage deficit compares success with a volume-matched reference under the same region-goal protocol.

\paragraph{Two controls.} The volume-matched model $\theta_{\mathcal V}$ removes the same number of transitions at random.
It is evaluated at the same training epoch as the hole model.
This comparison holds dataset size fixed and changes spatial coverage.
The success-matched model instead uses an earlier random-drop checkpoint.
Its frozen success equals the hole e3 model's success (Table~\ref{tab:adaptation_main}).
Online adaptation changes its success by $-2.4$ points.
Thus low initial success alone does not predict an adaptation benefit in this control.
The deficit below uses the volume-matched reference.

Writing $s^{\mathrm{fz}}(\theta)$ for frozen success on region goals,
\begin{equation}
  \mathcal{R}_{\mathrm{model}} = s^{\mathrm{fz}}(\theta_{\mathcal{V}}) - s^{\mathrm{fz}}(\theta_{\mathcal{H}}),
  \qquad
  \mathcal{R}^{\mathrm{post}}_{\mathrm{model}} = \mathcal{R}_{\mathrm{model}} - \Delta_{\mathrm{TTT}},
  \label{eq:modelregret}
\end{equation}
where $\Delta_{\mathrm{TTT}}$ is the frozen-to-online improvement on the hole
model. At the e3 checkpoint $\mathcal{R}_{\mathrm{model}} = 50.0$\,pts and
$\Delta_{\mathrm{TTT}} = 18.8$\,pts, leaving
$\mathcal{R}^{\mathrm{post}}_{\mathrm{model}} = 31.2$\,pts that test-time
training does not recover. At e5 the deficit is $39.2$ and the gain is $18.8$, leaving $20.4$
(Table~\ref{tab:regret}).

The reference is one volume-matched model.
Equation~\ref{eq:modelregret} therefore measures a descriptive success gap.
It does not bound the performance of other models or depth allocations.

\subsection{Stopping and Adaptation}

The factorial experiment tests whether stopping changes the effect of online adaptation.
Write the cell means
of the $2\times2$ factorial as $\mu_{a\theta}$ for allocation
$a \in \{\text{fixed}, \text{adaptive}\}$ and training
$\theta \in \{\text{frozen}, \text{TTT}\}$. The interaction contrast is
\[
  \iota = (\mu_{\mathrm{ada},\mathrm{TTT}} - \mu_{\mathrm{fix},\mathrm{TTT}})
        - (\mu_{\mathrm{ada},\mathrm{fz}} - \mu_{\mathrm{fix},\mathrm{fz}}),
\]
and an additive effect on success corresponds to $\iota=0$.
The mean interaction is $+0.4$ points.
Within-seed interactions are $\{-6,-4,+4,0,+8\}$ points.
The interaction changes sign across seeds.
The small mean does not establish independent effects.
Adaptive stopping lowers success in both training conditions.
Online adaptation improves success under both allocation rules.

\begin{table}[t]
\centering
\small
\caption{Stopping and online adaptation. Above, success relative to fixed-depth and coverage references. Below, the TwoRoom hole-e3 factorial uses seeds $\{42,7,123,5,99\}$ and 50 episodes per seed. Cells report mean success. The mean interaction is $+0.4$ pp, with within-seed values $\{-6,-4,+4,0,+8\}$ pp. The factorial and Table~\ref{tab:adaptation_main} are separate evaluations of the same labeled condition.}
\label{tab:regret}
\begin{minipage}[t]{\linewidth}
\centering
\setlength{\tabcolsep}{3.5pt}
\begin{tabular}{lcccc}
\toprule
 & \multicolumn{3}{c}{Success and fixed-depth gap} & Coverage \\
Env. & Best $d$ & $\deltaefe$ & $\Delta$ & $\mathcal{R}^{\mathrm{post}}_{\mathrm{model}}$ \\
\midrule
TwoRoom  & $90.0$@$30$ & $88.0$ & $+2.0$ & $\cdot$ \\
PushT    & $88.4$@$30$ & $87.2$ & $+1.2$ & $\cdot$ \\
Reacher  & $72.0$@$30$ & $74.7$ & $-2.7$ & $\cdot$ \\
Cube     & $67.3$@$10$ & $63.3$ & $+4.0$ & $\cdot$ \\
\midrule
hole (e3) & \multicolumn{3}{r}{$50.0 \to 31.2$ (TTT $+18.8$, $38\%$)} & $31.2$ \\
hole (e5) & \multicolumn{3}{r}{$39.2 \to 20.4$ (TTT $+18.8$, $48\%$)} & $20.4$ \\
\bottomrule
\end{tabular}
\end{minipage}
\par\medskip
\begin{minipage}[t]{\linewidth}
\centering
\setlength{\tabcolsep}{4pt}
\begin{tabular}{lcc}
\toprule
 & Fixed & Adaptive \\
\midrule
Frozen       & $37.2$ & $33.2$ \\
Online (TTT) & $55.2$ & $51.6$ \\
\midrule
TTT effect      & \multicolumn{2}{c}{$+18.2$} \\
Adaptive effect & \multicolumn{2}{c}{$-3.8$} \\
Interaction     & \multicolumn{2}{c}{$+0.4$} \\
\bottomrule
\end{tabular}
\end{minipage}
\end{table}

\subsection{Per-solve cost regret}

The logged elite-cost trajectory shows whether refinement has converged under the model objective.
For one optimizer call with halt iteration $\hat n$, define
\[
 r_{\mathrm{ps}} = c_{\hat n}-\min_{1\le t\le\hat n}c_t \ge 0,
\]
where $c_t$ is the recorded elite-mean cost. This measures deterioration of
that summary after its best recorded value. It is neither episode-success
regret nor the evaluated cost of the returned mean action sequence. A
monotonically decreasing cost trace has $r_{\mathrm{ps}}=0$, even when external
success decreases. The identical-cost construction in Proposition 2 permits this case.

The calculation uses the primary controller configuration.
Every optimizer call with a complete cost trajectory is included, regardless of episode success.
Calls with only endpoint costs cannot provide the trajectory minimum.
Table~\ref{tab:per_solve} counts these calls as missing.

For scale context we also compute
\[
 \widetilde r_{\mathrm{ps}}=
 \frac{r_{\mathrm{ps}}}{c_3-\min_{1\le t\le\hat n}c_t}
 \quad\text{when the denominator is positive},
\]
using the controller's minimum allowed depth. Calls with zero
denominator remain in the raw average and are excluded from the normalized
average. Most excluded calls have positive raw regret, so the normalized average describes a different subset of calls. The ratio can exceed one. Calculations use the logger's four-decimal
cost records and give equal weight to each eligible optimizer call.

\begin{table}[htbp]
\centering
\caption{Primary-controller elite-summary regret.
Only complete trajectories in the listed seed subsets contribute to the
cost averages. All calls are counted regardless of episode outcome. The normalized row uses iteration 3 as its reference and excludes eight zero denominators, seven with positive raw regret. The trace subset differs from the full-seed episode-success comparison.}
\label{tab:per_solve}
\small
\begin{tabular}{lcccc}
\toprule
 & TwoRoom & Reacher & PushT & Cube \\
\midrule
Seeds with full traces & $1,2,3,4$ & $1,2$ & $3,4$ & $1,2$ \\
Calls with full traces & $241$ & $156$ & $124$ & $142$ \\
Calls without full traces & $62$ & $81$ & $196$ & $65$ \\
Mean $r_{\mathrm{ps}}$ & $6.395$ & $2.723$ & $0.075$ & $0.042$ \\
Calls with positive denominator & $239$ & $151$ & $124$ & $141$ \\
Mean $\widetilde r_{\mathrm{ps}}$ & $0.178$ & $0.247$ & $0.007$ & $0.015$ \\
Zero-denominator calls & $2$ & $5$ & $0$ & $1$ \\
\bottomrule
\end{tabular}
\end{table}

Cost scales, seed subsets, and optimizer-call counts differ across environments.
The table describes internal convergence on the available complete trajectories.
It does not estimate a causal effect of stopping or a paired relationship with episode success.

\section{Online Adaptation}
\label{app:online}
\label{tab:online}
Online adaptation improves the hole model in independently trained samples, but the frozen disagreement penalty changes sign between them. Table~\ref{tab:adaptation_main} reports the second training sample. Table~\ref{tab:online_replication} adds the first sample and the remaining controls. The hole-model gain is positive at every evaluation seed in the first sample. The volume-matched and full-data models gain less.

The success-matched reference uses an earlier random-drop checkpoint with the same frozen success as the hole model. Its success falls after adaptation. The volume-matched references instead use the same training epochs as the hole models and have higher frozen success. The factorial experiment evaluates the hole condition separately from these coverage comparisons.

Reacher gives mixed results. Online adaptation lowers success at the earlier checkpoint in both training samples and gives a small gain at the later checkpoint. Its interaction with disagreement seeking also changes sign. Prediction error on collected transitions falls, but this improvement does not ensure better control.

Adaptation also helps some models trained without a coverage hole. Its benefit therefore depends on more than the presence of the hole. The comparison with shuffled-target updates tests whether pairing each collected transition with its correct target matters.

\begin{table}[htbp]
\centering
\caption{Additional online-adaptation results. Success and changes are percentages and percentage points, respectively. Paired changes are reported separately from rounded success means. Sample 1 uses three paired seeds per TwoRoom condition. Its hole-model gain is positive in all three. Sample 1 Reacher uses five paired seeds, with four negative changes at e3. The separate sample 2 volume-matched references have frozen success $87.2$ at e3 and $83.2$ at e5. On collected Reacher transitions, prediction error decreases from approximately $1.1$ to $0.63$ under adaptation, and anchor error stays flat.}
\label{tab:online_replication}
\small
\setlength{\tabcolsep}{5pt}
\begin{tabular}{llrrr}
\toprule
\multicolumn{2}{l}{TwoRoom, sample 1} & Frozen & Online & Change \\
\midrule
\multicolumn{2}{l}{Coverage hole} & $54.7$ & $69.3$ & $+14.7$ \\
\multicolumn{2}{l}{Volume control} & $66.0$ & $70.7$ & $+4.7$ \\
\multicolumn{2}{l}{Full data} & $74.0$ & $75.3$ & $+1.3$ \\
\midrule
\multicolumn{2}{l}{Frozen penalty contrast} & \multicolumn{2}{c}{Sample 1} & Sample 2 \\
\multicolumn{2}{l}{Coverage hole} & \multicolumn{2}{c}{$+20.0$} & $-4.0$ \\
\midrule
\multicolumn{2}{l}{Reacher} & \multicolumn{2}{c}{Online change} & Interaction \\
Sample 1 & e3 & \multicolumn{2}{c}{$-6.8$} & $+3.2$ \\
Sample 1 & e5 & \multicolumn{2}{c}{$+2.4$} & $-3.2$ \\
Sample 2 & e3 & \multicolumn{2}{c}{$-4.0$} & $\cdot$ \\
Sample 2 & e5 & \multicolumn{2}{c}{$+0.3$} & $\cdot$ \\
\midrule
\multicolumn{2}{l}{Controls without a hole} & \multicolumn{2}{c}{Frozen success} & Change \\
Full data & e1 & \multicolumn{2}{c}{$\cdot$} & $+9.6$ \\
Random drop & e1 & \multicolumn{2}{c}{$\cdot$} & $+6.8$ \\
Random drop & e2 & \multicolumn{2}{c}{$\cdot$} & $-2.4$ \\
Full data & e2 & \multicolumn{2}{c}{$70.8$} & $+5.6$ \\
\bottomrule
\end{tabular}
\end{table}

Larger penalty gains coincide with increases in probe error during each Reacher training run (Table~\ref{tab:probe_penalty}). The forward test is weaker. Its mean difference is positive, but the gain does not hold across seeds and it misses the prespecified criterion. The forward test therefore does not confirm that probe error predicts when the penalty helps.

\begin{table}[htbp]
\centering
\caption{Reacher probe error and frozen-penalty changes in success. Differences are percentage points. The sample 2 forward test at e2 has two positive seeds out of five and misses its prespecified criterion.}
\label{tab:probe_penalty}
\small
\begin{tabular}{llrr}
\toprule
Training sample & Checkpoint & Probe MSE & Penalty change \\
\midrule
1 & e3 & $0.194$ & $+6.8$ \\
2 & e5 & $0.122$ & $+8.0$ \\
2 & e8 & $0.035$ & $+2.0$ \\
2 & e2, forward test & $0.131$ & $+3.2$ \\
\bottomrule
\end{tabular}
\end{table}

\section{Stopping Rules}
\label{app:stopping}

\begin{table}[htbp]
\centering
\caption{Stopping-rule comparison. Success is the mean percentage across seeds. The second line in each adaptive cell gives mean depth. Each seed uses 50 episodes and cap 30. Seed sets and thresholds are specified below. Mean differences are descriptive where seed sets differ. Missing entries indicate unevaluated configurations.}
\label{tab:stopping}
\small
\setlength{\tabcolsep}{2pt}
\begin{tabular}{lcccc}
\toprule
Halting signal & TwoRoom & PushT & Cube & Reacher \\
\midrule
$\deltaefe$ EMA-mean (primary) & \shortstack{$88.0$\\@ $22.0$} & \shortstack{$87.2$\\@ $28.3$} & \shortstack{$63.3$\\@ $28.1$} & \shortstack{$74.7$\\@ $22.3$} \\
\addlinespace[2pt]
no-EMA (v1) & \shortstack{$87.3$\\@ $16.4$} & $\cdot$ & $\cdot$ & $\cdot$ \\
\addlinespace[2pt]
epistemic-aware (adaEFE) & \shortstack{$90.0$\\@ $24.9$} & \shortstack{$96.0$\\@ $29.8$} & \shortstack{$72.0$\\@ $29.2$} & \shortstack{$78.5$\\@ $26.7$} \\
\addlinespace[2pt]
best-elite cost stall & \shortstack{$89.3$\\@ $9.3$} & \shortstack{$87.3$\\@ $15.3$} & \shortstack{$63.3$\\@ $15.8$} & \shortstack{$74.0$\\@ $9.3$} \\
\addlinespace[2pt]
elite-variance stall & \shortstack{$88.0$\\@ $8.5$} & \shortstack{$77.3$\\@ $20.9$} & \shortstack{$66.0$\\@ $10.3$} & \shortstack{$76.7$\\@ $19.0$} \\
\addlinespace[2pt]
mean-change stall & \shortstack{$88.7$\\@ $22.5$} & \shortstack{$89.3$\\@ $17.9$} & \shortstack{$65.3$\\@ $24.7$} & \shortstack{$76.0$\\@ $26.2$} \\
\addlinespace[2pt]
elite-stability (fixed radius) & \multicolumn{4}{c}{depth $30.0$ at thresholds $\{0.6,0.8,0.9\}$ (seed 42)} \\
\addlinespace[2pt]
elite stability, $t=1.0$ & \shortstack{$86.0$\\@ $14.8$} & \shortstack{$90.0$\\@ $23.7$} & \shortstack{$70.0$\\@ $21.1$} & \shortstack{$64.0$\\@ $21.2$} \\
\midrule
best fixed depth & \shortstack{$90.0$\\$N=30$} & \shortstack{$88.4$\\$N=30$} & \shortstack{$67.3$\\$N=10$} & \shortstack{$72.0$\\$N=30$} \\
\bottomrule
\end{tabular}
\end{table}

\begin{figure}[htbp]
\centering
\includegraphics[width=\linewidth]{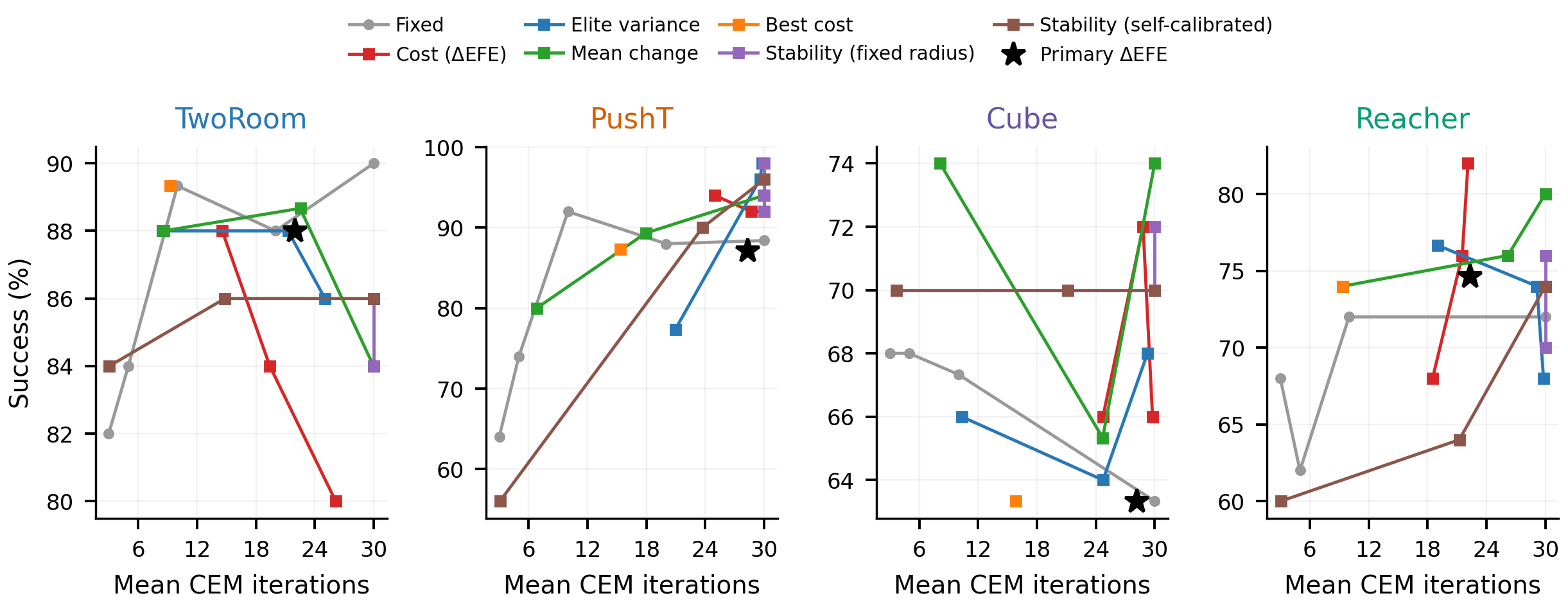}
\caption{Success and refinement depth for the stopping rules. Each signal is evaluated across its threshold grid. The gray line shows fixed-depth success and the black star marks the primary controller. Cost-based rules follow the plateau with different depth allocations. Geometric rules are sensitive to thresholds. No evaluated rule reliably recovers Cube's earlier success peak.}
\label{fig:pareto-stop}
\end{figure}

The primary controller uses seeds $\{42,1,2,3,4\}$ on TwoRoom and PushT.
It uses $\{42,1,2\}$ on Cube and Reacher.
Fixed-reference seeds are $\{1,2,3\}$ on TwoRoom, $\{42,1,2,3,4\}$ on PushT, and $\{42,1,2\}$ elsewhere.
Best-elite, variance, and mean-change rules use $\{42,1,2\}$.
Their shared thresholds are $10^{-2}$, $3\times10^{-2}$, and $10^{-1}$, respectively.
No-EMA uses $\{42,1,2\}$ on TwoRoom.
Epistemic-aware halting uses $\{42,1,2,3\}$ on Reacher and the elite-stability seed elsewhere.
Elite stability uses the single seed listed in Table~\ref{tab:stopping}.

All rules use the same solver, patience $p=2$, minimum depth $n_{\min}=3$, and candidate counts.
They differ in the statistic used to detect stalling.
Variance and mean-change rules monitor the CEM sampling distribution.
Best-elite stall monitors the minimum elite cost.
Elite stability measures the fraction of current elites within an action-space radius $0.5\bar\sigma\sqrt d$ of the previous elite set.
This radius-based rule never triggers at the tested thresholds.

A self-calibrated alternative compares cross-set nearest-neighbor distance with within-set spacing.
It stops when the median cross-set distance falls below $t$ times the median within-set distance.
The single-seed grid changes abruptly with $t$.
At $t=0.8$ the rule reaches the cap, whereas at $t=1.2$ it stops at the three-iteration minimum.
The intermediate setting $t=1.0$ gives the results in Table~\ref{tab:stopping}.
These measurements do not establish a stable geometric stopping signal.
Best-elite stall uses less depth than the primary rule, with nearby success means.
The evaluated rules do not reliably recover Cube's earlier success peak.

\subsection{Disagreement guards and detector results}
\label{app:guard}
A disagreement guard stops when the elite disagreement exceeds $\gamma$ times its running minimum.
We add this guard to the primary controller and vary $\gamma$ on Cube.
A shared setting is also evaluated on the three plateau tasks (Table~\ref{tab:guard}).
The tightest Cube guard stops earlier and recovers part of the lost success.
No tested setting reaches the earlier fixed-depth peak.
The shared guard also stops prematurely on PushT, where success declines substantially.
Thus the tested guard does not improve allocation consistently across tasks.

Four additional detectors use the recorded per-instance trajectories.
Two monitor disagreement rise.
Another compares imagined terminal cost with the realized outcome.
The fourth compares disagreement with its early-iteration level.
Their statistics and firing rates overlap across environments.
These detectors do not reliably separate Cube's over-optimization from the plateau tasks.
The result limits this detector family without ruling out externally informed stopping signals.

\begin{table}[htbp]
\centering
\caption{Disagreement guards. Mean success is followed by mean depth. Each cell uses three seeds and 50 episodes per seed. The guard stops when disagreement exceeds $\gamma$ times its running minimum. It supplements the primary controller. The plateau tasks use $\gamma=1.3$. At these seeds, the unguarded adaptive reference reaches $86$ on TwoRoom and $92$ to $98$ on PushT. Fixed-30 reaches $76$ on Reacher. Missing entries indicate unevaluated configurations.}
\label{tab:guard}
\small
\setlength{\tabcolsep}{3pt}
\begin{tabular}{lcccc}
\toprule
Rule & Cube & TwoRoom & PushT & Reacher \\
\midrule
fixed-10 (Cube peak)        & $71.3$ & $\cdot$ & $\cdot$ & $\cdot$ \\
fixed-30                    & $61.0$ & $\cdot$ & $\cdot$ & $\cdot$ \\
primary, unguarded& \shortstack{$67.0$\\@ $28.3$} & $\cdot$ & $\cdot$ & $\cdot$ \\
guard $\gamma{=}1.15$       & \shortstack{$68.7$\\@ $18.1$} & $\cdot$ & $\cdot$ & $\cdot$ \\
guard $\gamma{=}1.3$        & \shortstack{$64.0$\\@ $20.5$} & \shortstack{$82.0$\\@ $12.4$} & \shortstack{$69.3$\\@ $12.1$} & \shortstack{$78.7$\\@ $21.4$} \\
guard $\gamma{=}1.5$        & \shortstack{$66.7$\\@ $22.3$} & $\cdot$ & $\cdot$ & $\cdot$ \\
\bottomrule
\end{tabular}
\end{table}

\section{Normalized Weights and Sensitivity}
\label{app:scale}

Negative disagreement weights have different effects across tasks (Table~\ref{tab:beta_audited}). Reacher, PushT, and Cube gain at some weights, whereas every tested penalty lowers mean success on TwoRoom. Larger positive weights lower success on all tasks. The weight comparison reuses sampled starts and goals across seeds. The dropout generator is not explicitly seeded, so imagined trajectories are not fully paired.

Reacher's gains at stronger penalties are smaller in the larger seed sample than in the earlier evaluation. Its weaker penalties give both positive and negative differences. The results therefore do not support a uniformly beneficial negative-weight range.

Repeated evaluations also change the measured success of a fixed configuration. Reacher's initial evaluation was higher than every repetition. These repeated runs use the same evaluation seed and selected configurations, so they do not describe variation across training runs or all evaluation seeds.
\begin{table}[p]
\centering
\caption{Disagreement-weight comparison with 100 paired seeds and 50 episodes per run. Success is the mean percentage across seeds. $\Delta$ is the mean difference from the same-seed $\beta=0$ reference. Every configuration uses 30 CEM iterations and $K=4$ stochastic rollouts.}
\label{tab:beta_audited}
\small
\begin{tabular}{lrcc}
\toprule Env & $\beta$ & Success (\%) & $\Delta$ (pp) \\
\midrule
TwoRoom & -50 & $71.06$ & $-13.44$ \\
TwoRoom & -20 & $79.82$ & $-4.68$ \\
TwoRoom & -10 & $82.36$ & $-2.14$ \\
TwoRoom & -3 & $83.12$ & $-1.38$ \\
TwoRoom & 0 & $84.50$ & $\cdot$ \\
TwoRoom & +3 & $76.42$ & $-8.08$ \\
TwoRoom & +10 & $63.88$ & $-20.62$ \\
TwoRoom & +50 & $48.40$ & $-36.10$ \\
\midrule
PushT & -50 & $81.72$ & $-4.52$ \\
PushT & -20 & $86.40$ & $+0.16$ \\
PushT & -10 & $87.42$ & $+1.18$ \\
PushT & -3 & $87.66$ & $+1.42$ \\
PushT & 0 & $86.24$ & $\cdot$ \\
PushT & +3 & $70.12$ & $-16.12$ \\
PushT & +10 & $31.58$ & $-54.66$ \\
PushT & +50 & $8.18$ & $-78.06$ \\
\midrule
Reacher & -50 & $81.48$ & $+3.34$ \\
Reacher & -20 & $80.52$ & $+2.38$ \\
Reacher & -10 & $79.00$ & $+0.86$ \\
Reacher & -3 & $77.62$ & $-0.52$ \\
Reacher & 0 & $78.14$ & $\cdot$ \\
Reacher & +3 & $76.58$ & $-1.56$ \\
Reacher & +10 & $72.40$ & $-5.74$ \\
Reacher & +50 & $48.88$ & $-29.26$ \\
\midrule
Cube & -50 & $68.96$ & $+1.32$ \\
Cube & -20 & $69.08$ & $+1.44$ \\
Cube & -10 & $69.52$ & $+1.88$ \\
Cube & -3 & $68.66$ & $+1.02$ \\
Cube & 0 & $67.64$ & $\cdot$ \\
Cube & +3 & $67.58$ & $-0.06$ \\
Cube & +10 & $64.16$ & $-3.48$ \\
Cube & +50 & $54.66$ & $-12.98$ \\
\bottomrule
\end{tabular}
\end{table}

\begin{figure}[htbp]
\centering
\includegraphics[width=\linewidth]{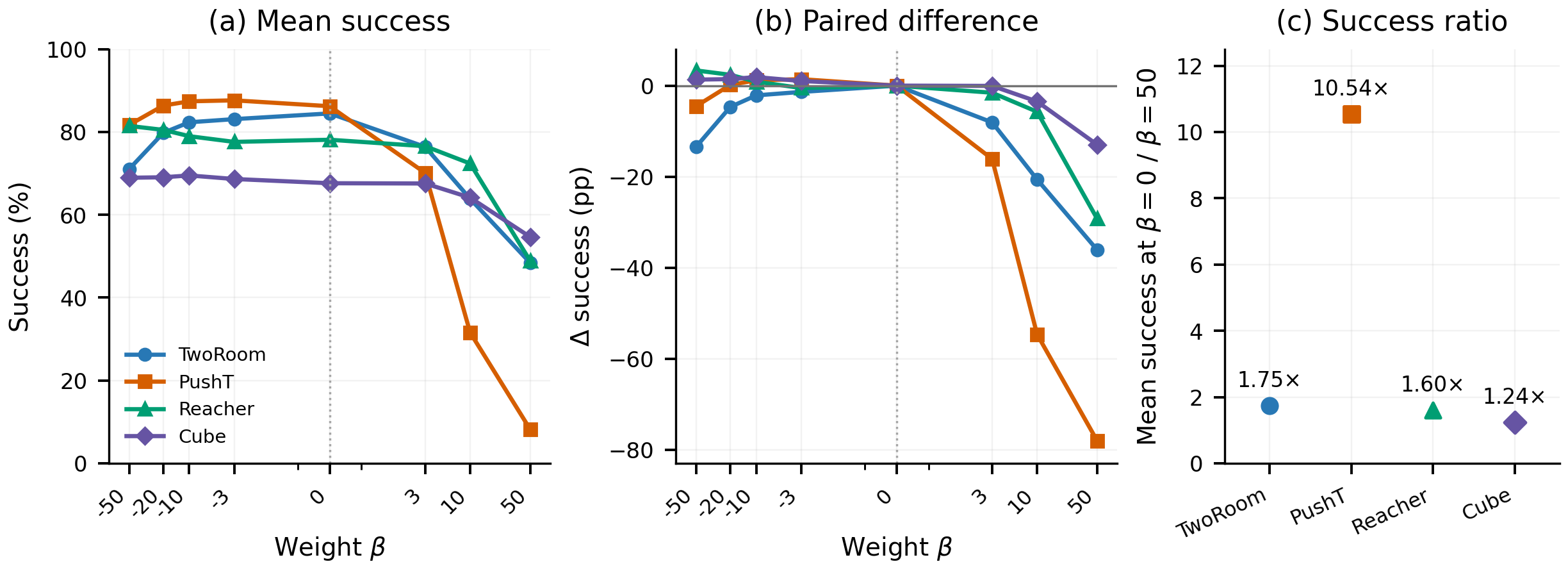}
\caption{Disagreement-weight results with 100 seeds per cell. (a) Mean success. (b) Mean paired differences from $\beta=0$. (c) Ratio of mean success at $\beta=0$ to $+50$, reaching $10.54\times$ on PushT. Each seed uses 50 episodes, giving 160,000 episode evaluations across the weight grid. Reacher's $\beta=-20,-50$ differences are $+2.38,+3.34$ pp, compared with $+4.4$ pp at each weight in the earlier 20-seed sample. In a separate set of repeated evaluations, Reacher success was $90\%$ initially and $80$ to $82\%$ in five repetitions.}
\label{fig:dosefull}
\end{figure}

\begin{figure}[htbp]
\centering
\includegraphics[width=\linewidth]{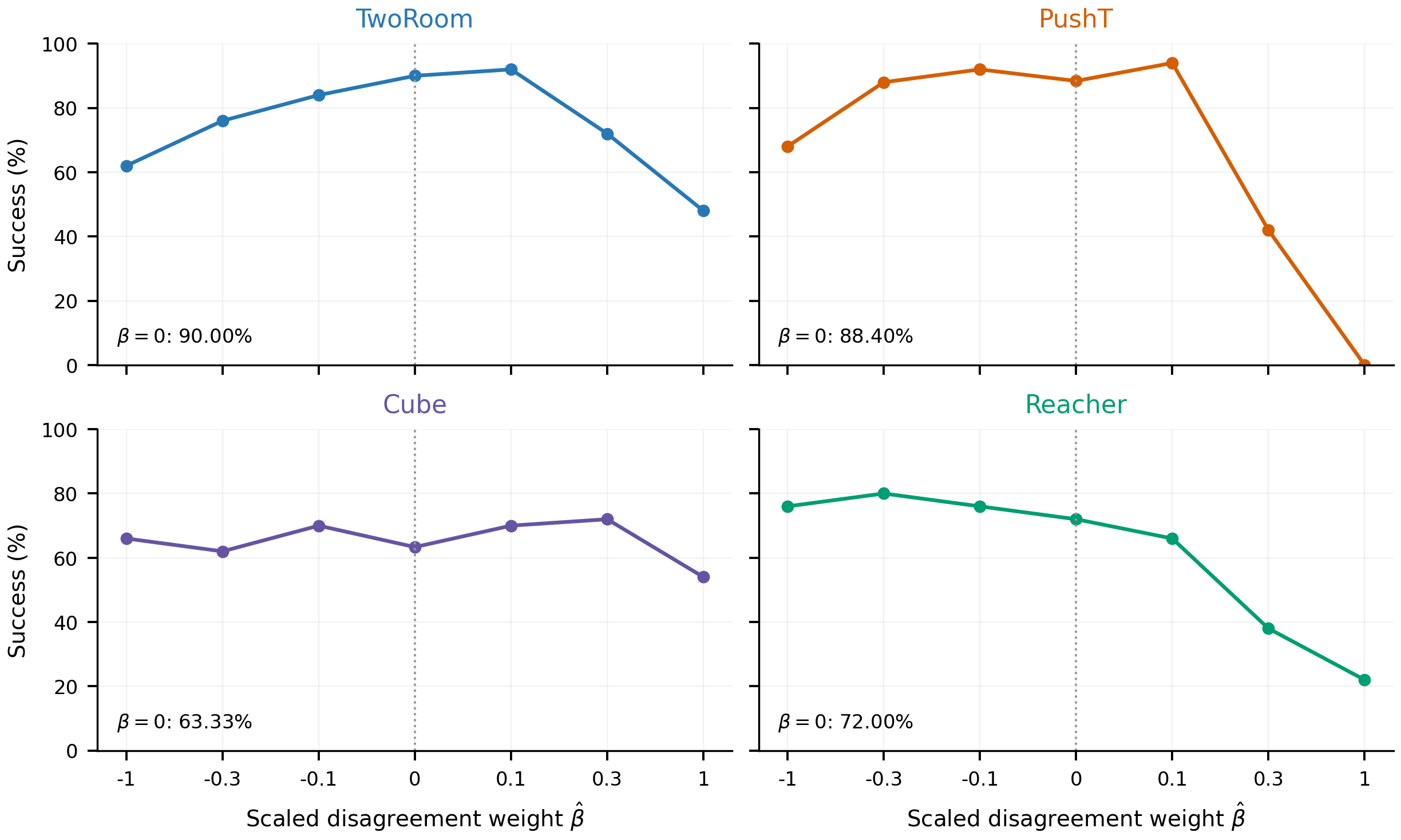}
\caption{Scale-matched dose response by environment. Seven normalized weights and 28 values use a common scale from $0$ to $100\%$. Baseline annotations give mean success. Nonzero-weight cells use a single seed. Disagreement magnitude is matched to the current candidate batch. The intermediate positive weight reduces success by $20$ to $52\%$ relative on three environments. The intermediate negative weight improves Reacher success by approximately $11\%$ relative.}
\label{fig:nbeta}
\end{figure}

We also scale disagreement by the current candidate-batch ratio of pragmatic and disagreement magnitudes.
The dimensionless weight is $\hat\beta\in\{-1,-0.3,-0.1,+0.1,+0.3,+1\}$ (Figure~\ref{fig:nbeta}).
The intermediate positive weight lowers success on most environments.
The largest positive weight gives zero observed success on PushT.
The intermediate negative weight improves mean success on Reacher.
Nonzero-weight cells use a single seed.
The sweep suggests that the positive-weight loss can persist after magnitude matching, but does not establish robustness across seeds.

Changing the ensemble size has little effect on Reacher success at the selected negative weight (Table~\ref{tab:proxysens}). The largest tested dropout rate lowers success. The cells share an evaluation seed, but the separate multi-seed baseline does not give a paired estimate of the gain. Dropout changes the imagined dynamics as well as the disagreement signal.

\begin{table}[htbp]
\centering
\caption{Proxy sensitivity on Reacher at $\hat\beta=-0.3$. The ensemble-size sweep fixes dropout at $p=0.1$. The dropout sweep fixes $K=4$. Each cell uses seed 42. The separate multi-seed $\beta=0$ baseline is $72.0$.}
\label{tab:proxysens}
\small
\begin{tabular}{lccc@{\hskip 16pt}ccc}
\toprule
 & $K{=}2$ & $K{=}4$ & $K{=}8$ & $p{=}0.05$ & $p{=}0.1$ & $p{=}0.2$ \\
\midrule
Success & $80$ & $80$ & $82$ & $76$ & $80$ & $64$ \\
\bottomrule
\end{tabular}
\end{table}

Candidate values in TD-MPC2 can share a large common offset.
Their magnitude can therefore exceed the variation that determines candidate ranking.
That variation also contracts as MPPI converges (Table~\ref{tab:spreadnorm}).
Magnitude matching can consequently overweight disagreement for either sign of $\hat\beta$.
We instead normalize by the per-task spread of candidate values.
The available columns show smaller return losses for negative than positive weights.
Their task subsets differ, so they do not provide a paired comparison across doses.
Candidate-value spread is a useful scale to consider when transporting the weight between planner families.

\begin{table}[htbp]
\centering
\caption{Spread-normalized weights on TD-MPC2. Raw $\beta$ equals $\hat\beta$ times per-task candidate-value spread divided by disagreement scale. Depth is fixed at 6. Return is a percentage of the $\beta=0$ baseline. Each column covers $n$ of 19 tasks, with different task subsets. Value magnitude is approximately $90\times$ candidate spread. The spread contracts approximately $70\times$ over six MPPI iterations. Dog-walk reaches $117.2\%$ at $\hat\beta=-1$ in one 10-episode evaluation.}
\label{tab:spreadnorm}
\small
\begin{tabular}{lcccc}
\toprule
$\hat\beta$ (spread-normalized) & $-0.3$ & $-1$ & $+0.3$ & $+1$ \\
\midrule
Return, \% of $\beta{=}0$ & $99.2\%$ & $96.5\%$ & $32.9\%$ & $5.9\%$ \\
$n$ tasks complete & $2$ & $10$ & $11$ & $10$ \\
\bottomrule
\end{tabular}
\end{table}

\end{document}